\documentclass{article}

\PassOptionsToPackage{numbers}{natbib}

\usepackage[preprint]{neurips_2025}

\usepackage[utf8]{inputenc} 
\usepackage[T1]{fontenc}    
\usepackage{hyperref}       
\usepackage{url}            
\usepackage{booktabs}       
\usepackage{amsfonts}       
\usepackage{nicefrac}       
\usepackage{microtype}      
\usepackage{xcolor}         
\usepackage{tabularx}
\usepackage{listings}

\usepackage{multirow}

\newcommand{\colhead}[1]{\multicolumn{1}{c}{\rotatebox{0}{\strut #1}}}
\usepackage{multirow}   
\usepackage{siunitx}    
\usepackage{adjustbox}  
\usepackage{subcaption} 
\usepackage{amsmath}
\usepackage{tcolorbox}
\tcbuselibrary{theorems,breakable}

\newtcbtheorem[]{myexample}{Example}%
{
  breakable,
  colback=green!5,
  colframe=green!35!black,
  fonttitle=\bfseries,
  top=2pt,         
  bottom=2pt,      
  before skip=4pt, 
  after skip=4pt   
}{th}

\newtcbtheorem[]{myanswer}{Answer}%
{
  breakable,
  colback=blue!5,
  colframe=blue!35!black,
  fonttitle=\bfseries,
  top=2pt,
  bottom=2pt,
  before skip=4pt,
  after skip=4pt
}{th}

\usepackage{wrapfig}
\usepackage{todonotes}

\title{PhySense: Principle-Based Physics Reasoning Benchmarking for Large Language Models}

\author{
Yinggan Xu$^{1\dagger}$ \quad Yue Liu$^{2\dagger}$ \quad Zhiqiang Gao$^{3}$ \quad
\vspace{5pt}
{\bf Changnan Peng$^{4}$ \quad Di Luo$^{5}$$^{*}$ } \\
$^1$Department of Computer Science, UCLA \\
$^2$Department of Physics, Caltech \\
$^3$Department of Physics, UC Berkeley \\
$^4$Department of Physics, MIT \\
$^5$Department of Electrical and Computer Engineering, UCLA \\
}

\begin{document}

\maketitle

\vspace{-1.5em}
\begin{abstract}
\vspace{-0.8em}
Large language models (LLMs) have rapidly advanced and are increasingly capable of tackling complex scientific problems, including those in physics. Despite this progress, current LLMs often fail to emulate the concise, principle-based reasoning characteristic of human experts, instead generating lengthy and opaque solutions. This discrepancy highlights a crucial gap in their ability to apply core physical principles for efficient and interpretable problem solving. To systematically investigate this limitation, we introduce \textit{PhySense}, a novel \textit{principle-based physics reasoning} benchmark designed to be easily solvable by experts using guiding principles, yet deceptively difficult for LLMs without principle-first reasoning. Our evaluation across multiple state-of-the-art LLMs and prompt types reveals a consistent failure to align with expert-like reasoning paths, providing insights for developing AI systems with efficient, robust and interpretable principle-based scientific reasoning.
\end{abstract}
\vspace{-1em}

\section{Introduction}

Large language models (LLMs) have emerged as powerful tools, profoundly impacting numerous aspects of scientific discovery \citep{zhang2024comprehensivesurveyscientificlarge, Romera-Paredes2024, lu2024aiscientistfullyautomated, shojaee2025llmsrscientificequationdiscovery}. Recent advancements in their reasoning capabilities have been particularly transformative, with notable applications in the domain of physics \citep{pan2025quantum, song2025llm, barman2025large,xu2025advancing}. Within physics, LLMs have demonstrated the ability to engage with problems ranging from those requiring real-world physical intuition \citep{geminiroboticsteam2025geminiroboticsbringingai} to complex theoretical challenges \citep{chung2025tp-bench}.

Despite these impressive strides, a critical challenge lies in ensuring that the reasoning processes of LLMs align with expert intuition and fundamental physical principles. Current LLMs tend to generate solutions with long-horizon reasoning pathways, which are opaque, convoluted, or divergent from the parsimonious and principle-driven thinking characteristic of human physicists. Such phenomena has also been identified as over-thinking \cite{sui2025stop}. In contrast, physicists master \textit{principle-based reasoning} with principle-driven problem solving and principle-based verification. Principle-driven problem solving is a forward process where fundamental principles simplify the problem-solving space, directly guiding towards a solution. Principle-based verification is a routine where physics principles establish criteria that a correct solution must meet, ensuring its validity. This divergence between LLMs and human physicists raises concerns about the \textit{efficiency, robustness and interpretability} of current LLMs for scientific reasoning, especially in a field where clarity, intuition and explainability of a solution is as crucial as the correctness of solution itself.

This work investigates LLMs' tendency to miss simple, intuitive solutions in physics problems that are apparent to human physicists. We posit that an incomplete grasp or misapplication of physical principles leads LLMs to unnecessarily complex reasoning, contrasting with human experts who leverage these fundamental ideas for elegant and efficient solutions (e.g., analyzing through symmetry instead of intricate numerical computation). This expert approach, which organizes knowledge around crystallized principles for efficient problem-solving, is well-documented in cognitive science \citep{LarkinJill1980EaNP, Chi1981, Sweller1988}. Emulating this in LLMs could foster more aligned, efficient, and interpretable reasoning, guiding them towards computationally leaner and conceptually sound `shorter paths.'

To systematically analyze this, we introduce \textit{PhySense}, a benchmark of 380 carefully curated physics problems designed to be straightforward for human physicists with core principles but appears to be challenging for LLMs to solve efficiently. In contrast to other physics reasoning benmark which focuses on reasoning on specific domain or challenging calculations, \textit{PhySense} focuses on short reasoning chains where single principles are crucial. Our findings aim to highlight the need for developing LLMs that are not only accurate, but also exhibit interpretable, robust reasoning aligned with fundamental physical principles. Our key contributions are summarized as follows:

\begin{figure}[t]
    \centering
    \includegraphics[width=\linewidth]{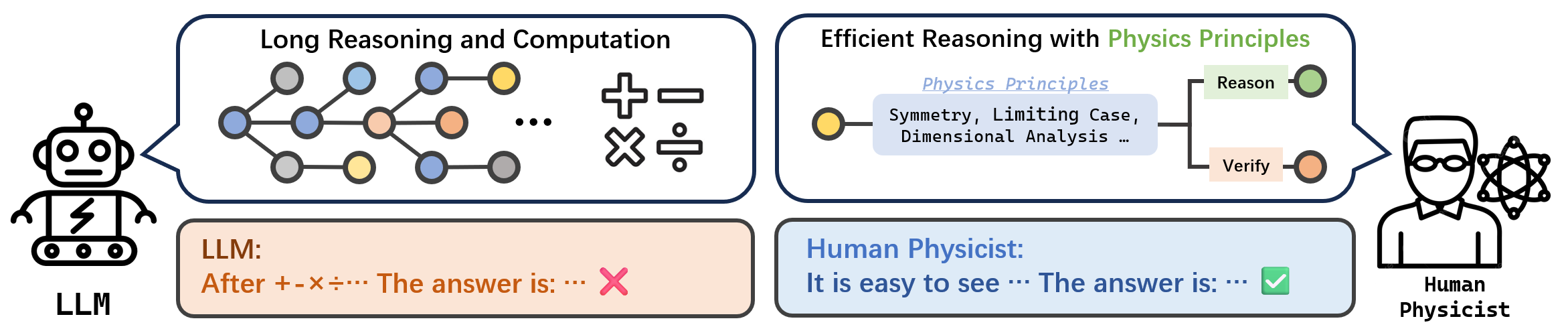}
    \caption{Illustrating how LLMs use lengthy, complex reasoning for physics problems intuitively straightforward to scientists applying core physical concepts.}
    \label{overall}
    \vspace{-15pt}
\end{figure}


    \quad \textbullet\; We introduce \textit{PhySense}, \textit{the first novel, human-curated principle-based physics reasoning benchmarking dataset} of over 380 problems that are straightforward for experts using fundamental principles but challenging to LLMs unless they adopt direct, principle-first reasoning shortcuts.
    
    \quad \textbullet\; We quantify not only whether an LLM arrives at the correct answer, but also how closely its reasoning cost matches with principle-based solutions via both accuracy and token efficiency metrics.
    
    \quad \textbullet\; We evaluate a range of state-of-the-art LLMs under zero-shot, hint, and no-computation prompts, uncovering LLMs' systematic lack of capability in applying principles and offering guidance for training LLMs toward more efficient, robust and interpretable principle-based physics reasoning.

\section{Related Work}\label{sec:related_work}
\vspace{-0.5em}
\textbf{Benchmarks for General Scientific Reasoning}\quad As LLMs are increasingly considered as important tools in scientific inquiry, understanding their true capabilities and limitations in scientific reasoning becomes paramount. Early benchmarks such as AI2 ARC \cite{clark2018arc}, MMLU \cite{hendrycks2020mmlu}, IconQA \cite{lu2021iconqa} and ScienceQA \cite{lu2022scienceQA} focused on general scientific context, surface-level reasoning, and basic factual knowledge. As model capabilities have grown, newer evaluations target deeper, multi-step problem solving and domain-specific expertise—either by repurposing advanced human exams and problem set (e.g., AGIEval \cite{zhong2023agieval}, JEEBench \cite{aroraetal2023jeebench}, SciBench \cite{wang2023scibench}) or by probing complex reasoning dimensions (e.g., MMLU-Pro \cite{wang2024mmlupro}, SciEval \cite{sun2024scieval}, TheoremQA \cite{chen2023theoremqa}), up to the extreme challenges posed by capstone-style assessments like Humanity’s Last Exam \cite{phan2025hle}. Some of the general science reasoning datasets like OlympicBench \cite{he2024olympiadbench} and OlympicAreana \cite{huang2024olympicarena} provides advanced physics problems but with limited scope.

\textbf{Benchmarks for Physics Reasoning}\quad The landscape of physics‐reasoning benchmarks for LLMs has rapidly evolved from primarily general problem sets to multifaceted collections that probe deeper conceptual, procedural, and physics-specific understanding. Efforts like PhyQA \cite{ding2023phyqa} and UGPhysics\cite{xu2025ugphysics} assemble thousands of structured introductory problems, while other benchmarks such as PhysBench \cite{qiu2025phybench} and PhysReason \cite{zhang2025physreason} introduce problems require longer reasoning steps. More research‐oriented suites like TP-Bench \cite{chung2025tp-bench}, CURIE \cite{cui2025curie} and multi-modal benchmarks like MM-PhyQA \cite{anand2024mm-phyqa} and domain specific benchmarks like FEABench \cite{mudur2025feabench} further pushes the understanding of LLM's physics capability with more research-oriented settings. Increased attention is also being directed towards fine-grained evaluation methodologies for the precise assessment of many-step reasoning including Expression Edit Distance (EED) Score \cite{qiu2025phybench}.

\textbf{Reasoning in LLMs and "Over-Thinking"}\quad Recent advances in LLMs, sometimes characterized by "slow thinking" capabilities demonstrated since models like GPT o1 \cite{openai2024reasoning}, have showcased stronger abilities in solving STEM problems. This improvement is often attributed to post-training techniques and reinforcement learning. Models like DeepSeek-R1\cite{guo2025deepseek}, Gemini-2.0-Flash-Thinking\cite{team2023gemini}, and versions of Claude \cite{anthropic2025claude37} and Qwen \cite{qwen2024qwq} have demonstrated enhanced reasoning. However, while these models can generate longer reasoning chains (i.e., use more tokens), this does not always equate to more efficient or accurate reasoning. The phenomenon of "over-thinking" \cite{sui2025stop}, where models may engage in unnecessarily complex or incorrect reasoning paths, remains a challenge.

\vspace{-7pt}
\section{Dataset Generation}
\vspace{-3pt}
\textit{``The universe is an enormous direct product of representations of symmetry groups.''}\\
\rightline{— Steven Weinberg, Nobel laureate in physics}

\textbf{Principle-based Reasoning} \quad Physics principles such as symmetries, conservation laws, and dimensional analysis remain cornerstones of modern physics research and problem solving. They not only simplify complex systems and reduce computational costs, but also illuminate the nature of various phenomena and provide a unified understanding across diverse contexts. Therefore, an LLM’s proficiency in applying these principles serves as a reliable gauge of its understanding of physics. Principle-based physics reasoning can (1) \textbf{efficiently yield the correct answer} (2) \textbf{robustly validate potential solutions} (3) \textbf{provide clear interpretability beyond calculation}. We demonstrate this with the following example.

\begin{myexample}{}{example1}
    A 5x5 square grid of nodes: $x \in \{0,1,2,3,4\}$, $y \in \{0,1,2,3,4\}$ connected by resistors $r$ between nearest neighbors. Connect node $V_{(0,0)} = 0$, node $V_{(4,4)} = V$, node $V_{(0,4)} = V/2$. Which of the following is true? 

    (a) $V_{(1,3)} = V/2$  (b) $V_{(2,2)} = V/2$ (c) $V_{(1,1)} = V/4$ (d) $V_{(3,3)} = 3V/4$ (e) $V_{(4,0)} = V/2$ 
\end{myexample}

\begin{myanswer}{}{answer by calculation}
\textbf{Answer by symmetry principle}:
\\
A trained physicist would notice the circuit together with added voltages has a reflection symmetry along the diagonal $x+y=4$. One can then deduce directly that (a,b,e) is correct.

\vspace{5pt}

\textbf{Answer by explicit calculation}:
\\
Without using symmetries, one has to solve Kirchhoff equations for the whole system (22 unknown voltages), 
\begin{equation}
    \begin{aligned}
        &3V_{0,1} - V_{1,1} - V_{0,2} = 0, \  3V_{0,2} - V_{1,2} - V_{0,1} - V_{0,3} = 0, \  3V_{0,3}-V_{1,3}-V_{0,2} = V/2\\
        & 3 V_{1,0}-V_{2,0}-V_{1,1}=0, \ 3 V_{2,0}-V_{1,0}-V_{3,0}-V_{2,1}=0, \  3 V_{3,0}-V_{2,0}-V_{4,0}-V_{3,1}=0 \\
        & 2 V_{4,0}-V_{3,0}-V_{4,1}=0, \ 3 V_{4,1}-V_{4,0}-V_{4,2}-V_{3,1}=0, \ 3 V_{4,2}-V_{4,1}-V_{4,3}-V_{3,2}=0\\
        & 3 V_{4,3}-V_{4,2}-V_{3,3}=V, \ 3 V_{1,4}-V_{2,4}-V_{1,3}=V/2, \ 3 V_{2,4}-V_{1,4}-V_{3,4}-V_{2,3}=0\\
        & 3 V_{3,4}-V_{2,4}-V_{3,3}=V, \ 4 V_{i, j}-V_{i-1, j}-V_{i+1, j}-V_{i, j-1}-V_{i, j+1}=0 \text{~for~} 1\leq i,j\leq 3.\\ \nonumber
    \end{aligned}
\end{equation}
Solving all the equations above numerically, one gets $V_{1,3}=V_{2,2}=V_{4,0}=V/2$, $V_{3,3}\approx 0.6702V$, $V_{1,1}\approx0.3298V$. Thus the answer is (a,b,e). Clearly, this ``standard'' approach is much more complicated than using the symmetry principle.
\end{myanswer}

\begin{wrapfigure}{O}{0.\textwidth}
  \centering
  \includegraphics[width=0.47\textwidth]{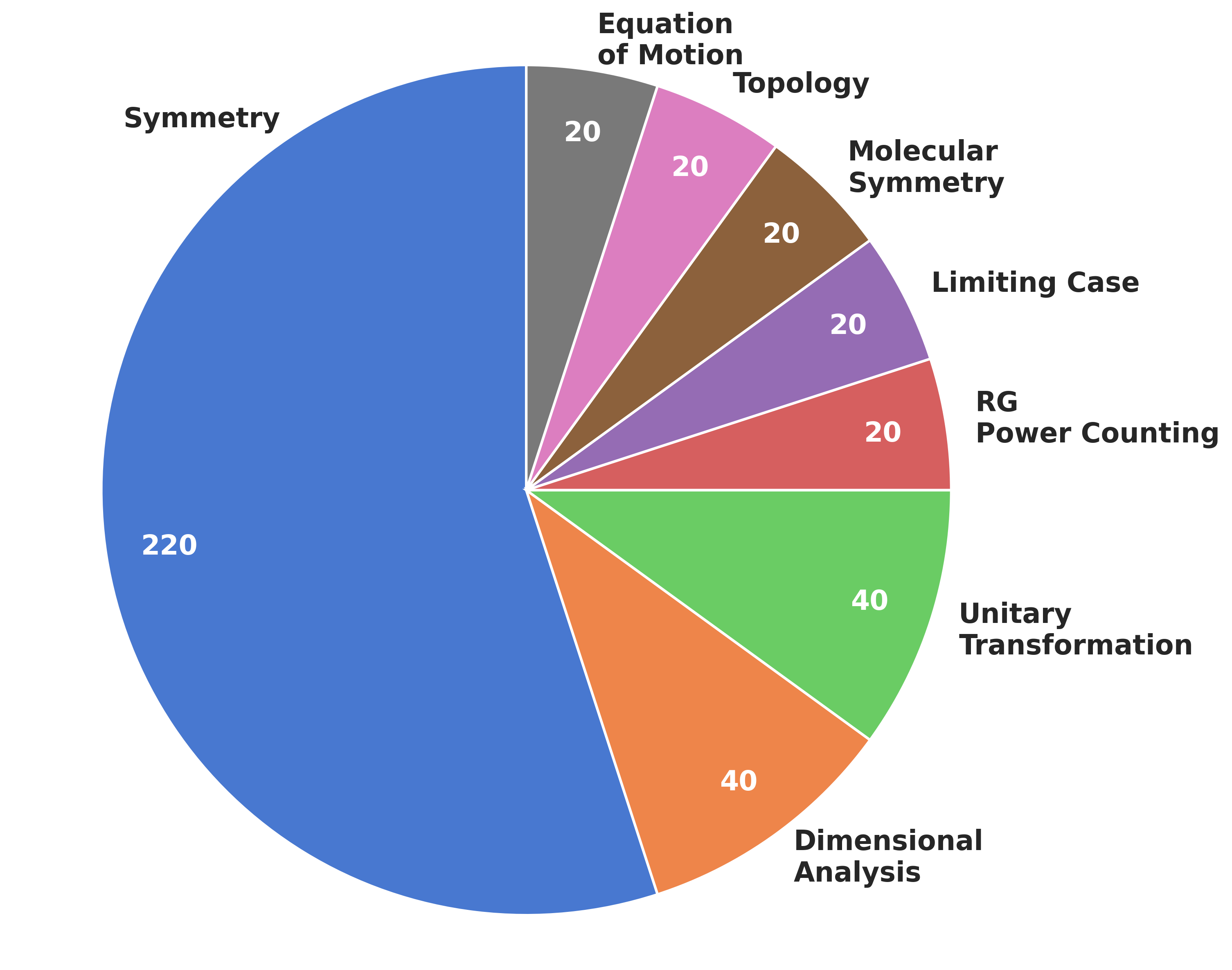}
  \caption{Distribution of physics principles in the dataset.}\label{fig:principle_pie}
  \vspace{-17pt}
\end{wrapfigure}

Despite the power of physical principles, existing benchmarks (see e.g. Sec.~\ref{sec:related_work}), while challenging, do not evaluate whether LLMs truly apply these principles. Do LLMs genuinely understand physics, or are they merely leveraging greater computational power than humans? To address this gap, we have developed a new problem set of 380 physics questions spanning electricity and magnetism, electric circuits, quantum spin/fermion chains, quantum dynamics, topological insulators, the renormalization group, and conformal field theory. These problems are crafted according to the following criteria.

\textbf{Principle-based physics reasoning}\quad A key feature of \textit{PhySense} is its design to test LLMs' understanding on fundamental principles and capability on principle-based reasoning. Our dataset is different than previous physics reasoning dataset, since we do not aim to test LLMs' knowledge in a specific domain or cabability of reasoning with long calculation. While our problems may be challenging or could be solved with lengthy calculation, we design the problems to be solved easily using physics principle reasoning.

\textbf{Novel problems from human experts}\quad Although the underlying concepts in our problem set are widely available online, we have crafted entirely new questions with physicists from top universities that cannot be found elsewhere, ensuring that LLMs have not been exposed to similar problems. This novelty is essential for testing an LLM’s ability to generalize the application of physics principles.

\textbf{A wide range of difficulties}\quad The problems span difficulty levels from undergraduate through graduate and research-level, yet none requires advanced mathematical techniques, complicated integrals, or large-scale numerical computations. This ensures we evaluate how well LLMs can think like physicists --- using fundamental physical principles to understand problems --- rather than merely assessing raw computational capability. We also annotate each problem with a difficulty rating (as judged by humans) for subsequent analysis.

\textbf{Conciseness for evaluation}\quad Every problem is stated and solved entirely through textual description and derivation. The physical setups are simple to describe, minimizing the risk of misinterpretation by LLMs. To eliminate ambiguity in the outputs, each question offers either multiple-choice options or expects a concise numerical answer.

\subsection{Physical principles and models}

Following the criteria above, we evaluate the LLM's understanding and correct application of several fundamental yet powerful principles in both classical and quantum physics. To do this, we design 19 distinct problem models 

\textbf{Symmetry}\quad Spatial symmetries can be leveraged to identify points where complicated integrals vanish. To evaluate this, we construct problem sets involving two-dimensional and three-dimensional electric (or magnetic) fields generated by symmetric charge (or current) distributions. These problems are categorized into the following models, each with an abbreviation:

\quad \textbullet\; \textit{2D electric field (2DEF), 2D electric field on a lattice (2DEFL), 3D electric field (3DEF), 2D magnetic field (2DBF), 3D magnetic field (3DBF)}

    We also devise problems that leverage symmetries to determine voltages of certain nodes in finite and infinite circuits:

    \quad \textbullet\;\textit{Infinite resistive lattices (InfRes), Circuits on a square lattice (SqGrd), Circuits on other lattices (OthGrd)}

    The symmetry of molecules can determine the solubility in solvents, which leads to another model of problems:

    \quad \textbullet\;\textit{Solubility comparison (Solub)}
        
    Moreover, symmetries impose constraints on correlation functions in quantum many-body physics and statistical mechanics. We have also developed problems involving quantum spin and fermion chains and their dynamical variants, to test $\mathbb{Z}_2$, $U(1)$, and time-reversal symmetries alongside spatial symmetries such as translation and reflection:

    \quad \textbullet\;\textit{Quantum spin chains (Qspin), Fermionic chains (Ferm), Quantum dynamics with symmetry and conservation laws (DynCon)}

\textbf{Dimensional analysis}\quad Dimensional analysis is a powerful tool in uncovering possible relations between different physical quantities. Not only is it widely used in the context of thermodynamics, fluid mechanics, etc., its applications also extend to quantum mechanics as well. We design problems in two areas: (a) applying the $\Pi$ theorem in fluid and quantum mechanics, and (b) using power counting to determine relevance in the renormalization group. This yields the following problem models:

\quad \textbullet\;\textit{Dimensional analysis using $\Pi$ theorem, where we focus on testing LLM's ability to compute dimensions in arguments of functions such as $\sin, \log$, etc. (DimLS), Dimensional analysis with artificial irrelevant perturbations (WrdH), Power-counting in renormalization group analysis (RGPow)}

\textbf{Limiting case}\quad Irrelevant perturbations in physical problems can be omitted to simplify the physical model. To test the LLM’s ability to do so, we introduce perturbations into Model (WrdH) above and evaluate whether it correctly ignores the higher-order terms.
    
\textbf{Conservation law}\quad Conservation law plays a crucial role in quantum field theory. Especially in free fermion conformal field theories, equation of motion, together with the fermionic statistics, provides a powerful tool to determine whether an operator is primary, descendant, or merely vanishing.

\quad \textbullet\;\textit{Operator properties in conformal field theories (CFTOp)}
        
\textbf{Topology}\quad Topological phases of matter is a central topic in modern condensed matter physics. It typically exhibits gapless edge spectrum, and sensitive to the boundary condition of the system. We design problems to evaluate if LLMs can understand the stability of symmetry-protected topological phases from the edge spectrum perspective:

\quad \textbullet\;\textit{Edge spectrum in topological insulators (GpEdg)}
        
    We also compose problems in counting the ground state degeneracy of (generalized) spin chain with antiperiodic or periodic boundary condition. In particular, in these problems, applying finite-depth local unitary circuits, which does not alter the topological property including the ground state degeneracy, greatly simplifies the calculation.

    \quad \textbullet\;\textit{Ground state degeneracy of spin chains (GSDeq), Ground state degeneracy of generalized spin chains (GSDGen)}

\section{Experiments}
\vspace{-1em}
This section details the experiments conducted to evaluate the scientific problem-solving capabilities of LLMs. We begin by outlining the experimental setup, including the models tested and the prompting strategies employed to simulate scientific reasoning scenarios.

\subsection{Experiment Setup}
We evaluated seven unimodal LLMs on our benchmark. These included four \textbf{reasoning models} optimized for reasoning: GPT o4-mini-high \cite{openai2025o3o4}, Claude Sonnet 3.7 Thinking \cite{anthropic2025claude37}, Gemini 2.5 Pro \cite{GoogleDeepMind_GeminiUpdate_2025}, and DeepSeek R1 \cite{guo2025deepseek}. Additionally, we tested three regular \textbf{non-reasoning models}: GPT 4.1 \cite{OpenAI_GPT4.1_2025}, Claude 3.7 Sonnet \cite{anthropic2025claude}, and DeepSeek V3 \cite{deepseekai2025deepseekv3technicalreport}. For all the models, we use the API-based services with default hyperparameter setting. We utilized three common prompting strategies in scientific applications to test LLMs:

\textbf{Zeroshot Prompting}\quad In this setting, models received only the problem statement and format requirements, with no illustrative examples or hints regarding applicable physics principles. This strategy assesses the models' inherent problem-solving capabilities, relying on their existing background knowledge and reasoning abilities. The flexibility of zero-shot prompting makes it highly relevant to scientific discovery, testing LLMs' awareness and capability of principle-based reasoning. 

\textbf{Hint Prompting}\quad LLMs may occasionally fail to apply appropriate physical principles, either due to a lack of awareness of the principle or its incorrect application. In this setting, models were provided with hints regarding the relevant physical principles to apply, typically based on the problem's category or core concepts. This approach investigates the models' ability to utilize explicit guidance in their reasoning process.

\textbf{No Computation Prompting}\quad We observed that some LLMs, particularly reasoning models, may default to complex computational approaches, potentially obscuring the application of fundamental principles and leading to overly convoluted solutions. In this setting, models were explicitly instructed to avoid complex computations and instead focus on principle-based reasoning. This strategy aims to assess if LLMs can prioritize simpler, principle-driven solutions when they are not allowed to do computation.

\vspace{-7pt}
\subsection{Metrics}
\vspace{-7pt}

We employ two primary metrics for evaluation: \textbf{accuracy} and \textbf{token usage}. For accuracy, LLMs were instructed to provide their final answer within a boxed environment for automated extraction and comparison against ground truth solutions. The problems fall into two categories with different evaluation implementation: (1) numerical: Answers are compared to the ground truth allowing for a 5\% tolerance. (2) multiple choice: The selected option must exactly match the correct choice. For token usage, we record the total number of tokens produced during the generation of the solution for each problem and model. This is a crucial metric that provides insight into the computational cost associated with each model's problem-solving process, and reflects how much principle-based reasoning each LLM acquires.

\vspace{-7pt}
\section{Results}
\vspace{-7pt}

\begin{figure}[t]
  \centering
  \begin{subfigure}[b]{0.48\linewidth}
    \centering
    \includegraphics[width=\linewidth]{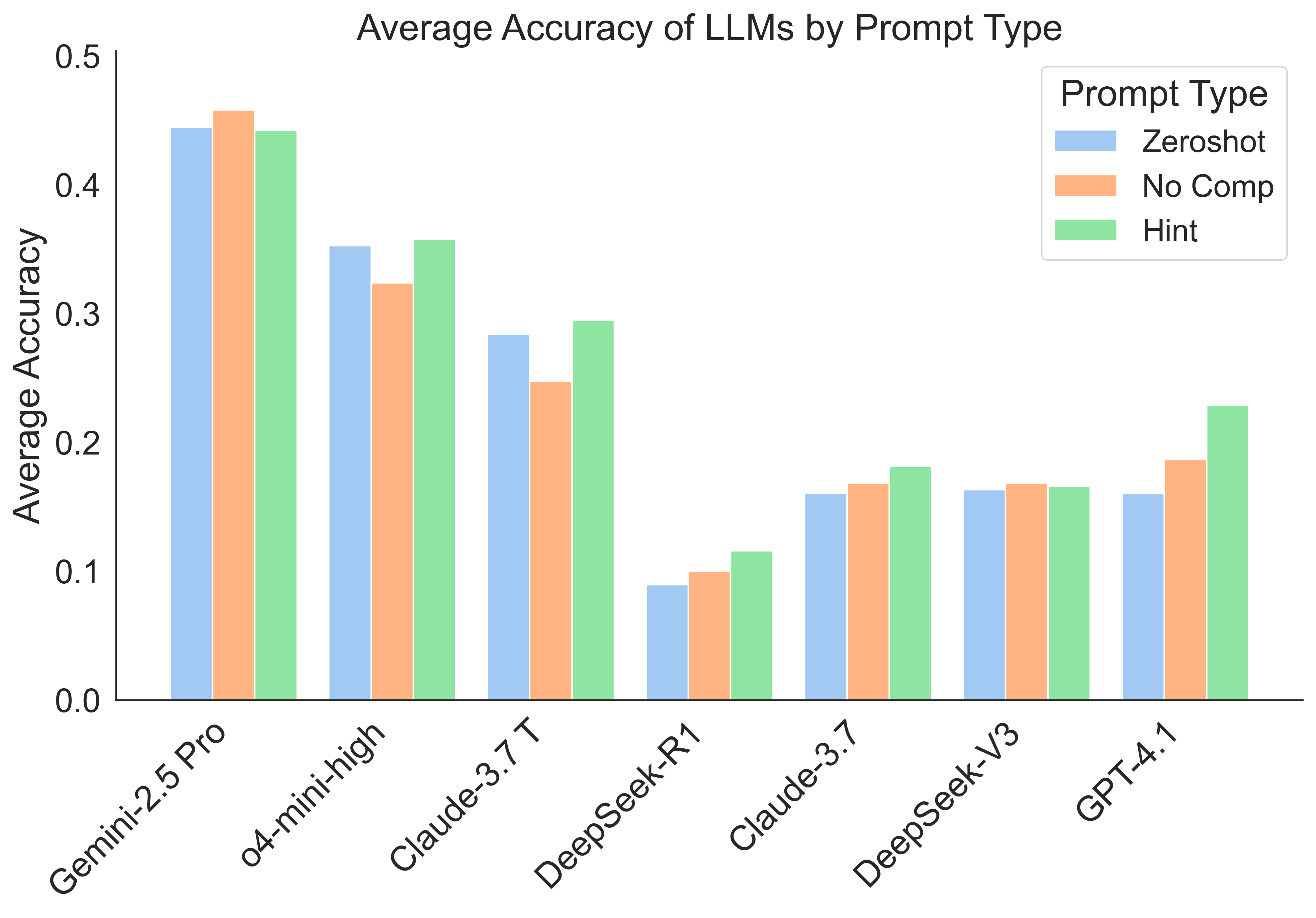}
    \caption{Average accuracy across models.}
    \label{fig:average_accuracy}
  \end{subfigure}
  \hfill 
  \begin{subfigure}[b]{0.48\linewidth}
    \centering
    \includegraphics[width=\linewidth]{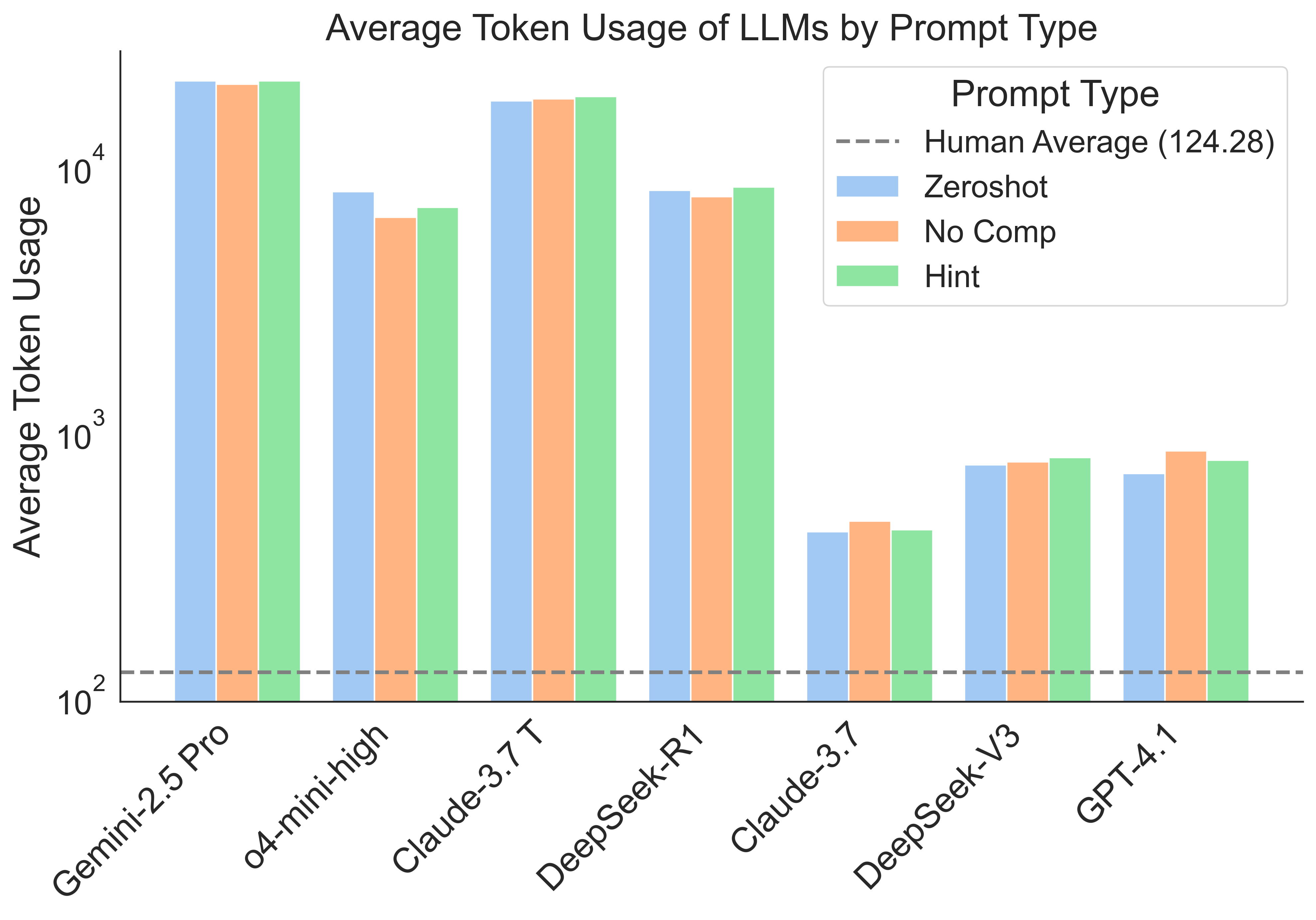}
    \caption{Average token usage across models.}
    \label{fig:average_token}
  \end{subfigure}
  \caption{Average accuracy and token usage for different models.}
  \label{fig:combined_average_plots}
\vspace{-20pt}
\end{figure}

In this section, we report the benchmarking results and present our primary observations regarding the performance of various LLMs on PhySense.
\vspace{-5pt}
\subsection{Reasoning Accuracy}
\vspace{-5pt}

\begin{wrapfigure}{O}{0.\textwidth}
  \centering
  \includegraphics[width=0.45\textwidth]{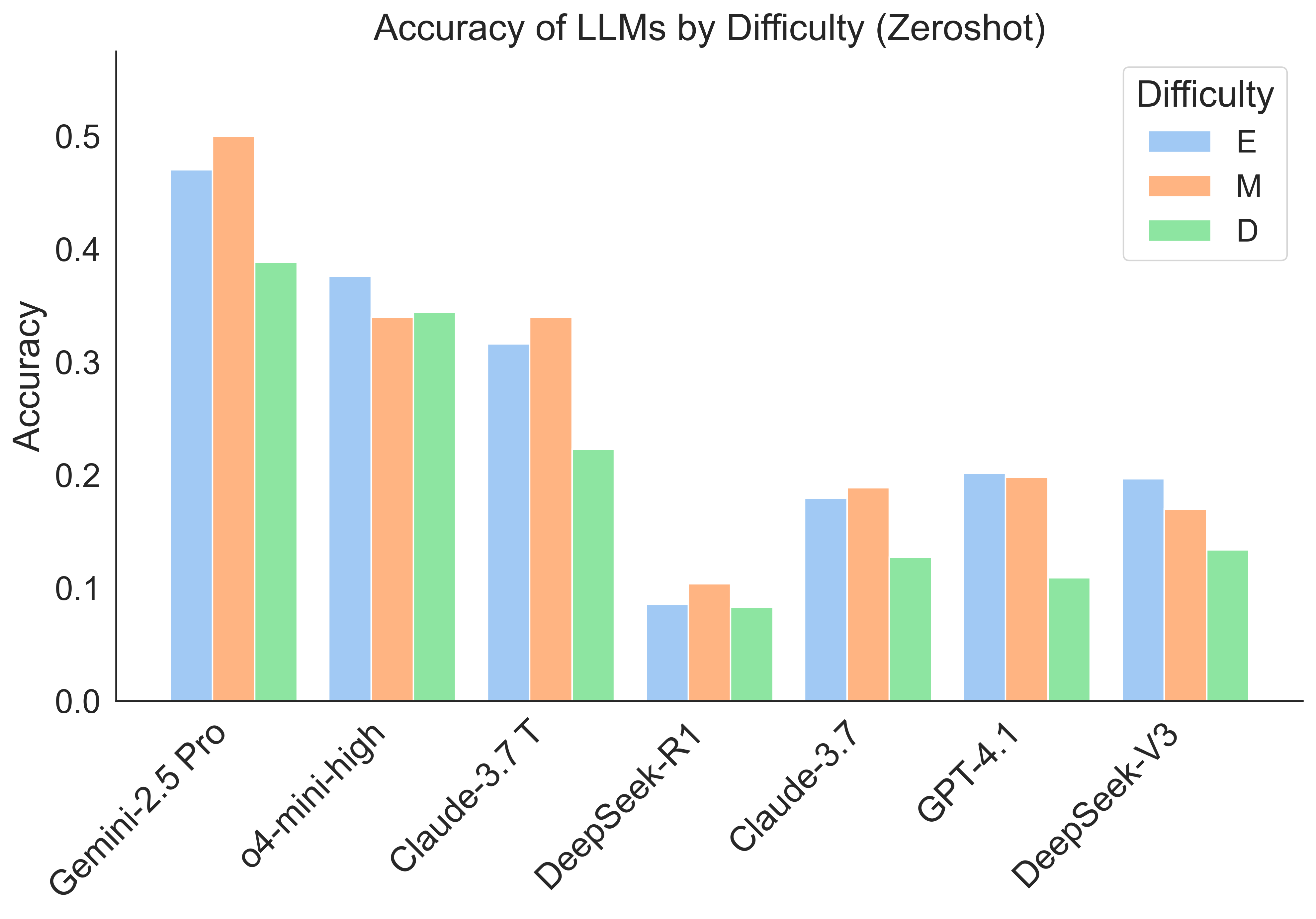}
  \caption{Zeroshot accuracy of LLMs under three difficulties.}\label{fig:principle_pie}
  \vspace{-10pt}
  \label{difficulty_acc}
\end{wrapfigure}

We report the model performance in terms of accuracy score for each section and an average accuracy over all problems.  We quantify model performance using accuracy percentage, calculated for each distinct problem category within our benchmark, alongside an overall average accuracy across all problems. This accuracy reflects the proportion of problems correctly solved by each model according to our evaluation protocol. The accuracy results are compiled in Table \ref{tab:llm_accuracy_percent_split_thinking}. To provide a clearer visual summary of the overall performance trends, we present histograms illustrating the distribution of average accuracy scores across the cohort of tested models in Figure \ref{fig:average_accuracy}. To further assess LLM alignment with human physicist problem-solving, problems were categorized by human-judged difficulty (easy, medium, difficult). Figure \ref{difficulty_acc} shows each model's average zero-shot accuracy across these levels. While reasoning models achieve better performance than non-reasoning models on average, all LLMs' performances are not satisfactory, reflecting their incapability of mastering principle-based reasoning.

\vspace{-7pt}
\subsection{Reasoning Token Efficiency}
\vspace{-7pt}
In parallel to accuracy, we report the average number of completion tokens produced by the models for generating solutions, both for individual sections and on average. This token usage metric provides an indication of the computational resources and reasoning complexity associated with each model's problem-solving attempts. The token usages are compiled in Table \ref{tab:llm_token_usage_split_thinking}. Figure \ref{fig:average_token} offers a comparative overview of the token utilization patterns. Reasoning models, due to reaoning mechanisms, consume about ten times more tokens ($\sim 10^4$) than non-reasoning models ($\sim 10^3$). In stark contrast, human physicists demonstrate far greater efficiency, often solving the same problems using about a hundered times fewer tokens ($\sim 10^2$) than reasoning models. It indicates a huge gap between LLMs and human experts on efficient principle-based reasoning.

\begin{table}[!ht]
    \centering
    \sisetup{round-mode=places, round-precision=1, table-auto-round}
    \small 
    \caption{LLM accuracy scores (as percentages) for reasoning models. The first subtable shows accuracy for the first 10 problem sets, and the second subtable shows accuracy for the remaining 9 problem sets and the overall average (AVG). The best accuracy of each section is marked in bold font. The full result, including non-reasoning models, is in the Appendix.}
    \begin{subtable}{\linewidth}
        \centering
        \label{subtab:accuracy_percent_problems_1_10_thinking_v2}
        \begin{adjustbox}{max width=\textwidth, center}
        \begin{tabular}{@{}ll *{10}{S[table-format=3.1, table-space-text-post=\%]}@{}}
        \toprule
        \textbf{Model} & \textbf{Prompt} &
        \colhead{RGPow} & \colhead{SqGrd} & \colhead{QSpin} & \colhead{CFTOp} & \colhead{3DBF} &
        \colhead{GSDGen} & \colhead{WrdH} & \colhead{Ferm} & \colhead{DynCon} & \colhead{3DEF} \\
        \cmidrule(lr){3-12}

        \multirow{3}{*}{DeepSeek R1}
        & Hint      & \textbf{25.0} & 5.0 & 0.0 & 0.0 & 5.0 & 0.0 & 55.0 & 0.0 & 0.0 & 5.0 \\
        & No Comp     & 15.0 & 5.0 & 5.0 & 0.0 & 5.0 & 0.0 & 40.0 & 5.0 & 0.0 & 0.0 \\
        & Zeroshot  & 10.0 & 0.0 & 10.0 & 5.0 & 10.0 & 0.0 & 30.0 & 5.0 & 0.0 & 5.0 \\
        \midrule
        \multirow{3}{*}{\shortstack[l]{Claude 3.7 \\ Sonnet \\ (Thinking)}}
        & Hint      & 5.0 & 30.0 & 35.0 & 30.0 & 40.0 & 0.0 & 65.0 & 20.0 & 5.0 & 50.0 \\
        & No Comp     & 10.0 & 25.0 & 30.0 & 20.0 & 45.0 & 0.0 & 25.0 & 10.0 & 0.0 & 30.0 \\
        & Zeroshot  & 10.0 & 45.0 & 35.0 & \textbf{35.0} & 35.0 & 0.0 & 30.0 & 25.0 & 5.0 & 30.0 \\
        \midrule
        \multirow{3}{*}{O4-Mini-High}
        & Hint      & 5.0 & 20.0 & 45.0 & 25.0 & 35.0 & 15.0 & 70.0 & \textbf{50.0} & 0.0 & 40.0 \\
        & No Comp     & \textbf{25.0} & 10.0 & 45.0 & 20.0 & 45.0 & 20.0 & 50.0 & 35.0 & 5.0 & 50.0 \\
        & Zeroshot  & 15.0 & 15.0 & 35.0 & 15.0 & 45.0 & 20.0 & 80.0 & 15.0 & 10.0 & 65.0 \\
        \midrule
        \multirow{3}{*}{\shortstack[l]{Gemini 2.5 Pro \\ (Preview)}}
        & Hint      & 10.0 & \textbf{50.0} & 65.0 & 25.0 &\textbf{50.0} & 5.0 & \textbf{100.0} & 30.0 & 25.0 & 65.0 \\
        & No Comp     & 20.0 & 40.0 & 65.0 & 25.0 & \textbf{50.0} & \textbf{25.0} & \textbf{100.0} & 25.0 & 20.0 & \textbf{70.0} \\
        & Zeroshot  & 10.0 & 35.0 & \textbf{70.0} & 25.0 & 40.0 & 15.0 & 95.0 & 25.0 & \textbf{30.0} & 50.0 \\
        \bottomrule
        \end{tabular}
        \end{adjustbox}
    \end{subtable}

    \vspace{0.3em} 

    \begin{subtable}{\linewidth}
        \centering
        \label{subtab:accuracy_percent_problems_11_avg_thinking_v2}
        \begin{adjustbox}{max width=\textwidth, center}
        \begin{tabular}{@{}ll *{10}{S[table-format=3.1, table-space-text-post=\%]}@{}}
        \toprule
        \textbf{Model} & \textbf{Prompt} &
        \colhead{DimLS} & \colhead{GpEdg} & \colhead{GSDeg} & \colhead{Solub} & \colhead{2DEF} &
        \colhead{2DEFL} & \colhead{OthGrd} & \colhead{2DBF} & \colhead{InfRes} & \colhead{AVG} \\
        \cmidrule(lr){3-12}

        \multirow{3}{*}{DeepSeek R1}
        & Hint      & 15.0 & 5.0 & 30.0 & 0.0 & 10.0 & 5.0 & 10.0 & 10.0 & 40.0 & 11.5789474 \\
        & No Comp     & 0.0 & 0.0 & 20.0 & 0.0 & 5.0 & 0.0 & 15.0 & 5.0 & 70.0 & 10.0000000 \\ 
        & Zeroshot  & 5.0 & 10.0 & 30.0 & 0.0 & 5.0 & 0.0 & 0.0 & 0.0 & 45.0 & 8.9473684 \\
        \midrule
        \multirow{3}{*}{\shortstack[l]{Claude 3.7 \\ Sonnet \\ (Thinking)}}
        & Hint      & 45.0 & 5.0 & 15.0 & 60.0 & 40.0 & 15.0 & 35.0 & 15.0 & 50.0 & 29.4736842 \\
        & No Comp     & 45.0 & 5.0 & 20.0 & 65.0 & 45.0 & 15.0 & 40.0 & 5.0 & 35.0 & 24.7368421 \\
        & Zeroshot  & 50.0 & 5.0 & 30.0 & 65.0 & 35.0 & 15.0 & 40.0 & 15.0 & 35.0 & 28.4210526 \\
        \midrule
        \multirow{3}{*}{O4-Mini-High}
        & Hint      & \textbf{55.0} & 10.0 & \textbf{45.0} & 45.0 & 45.0 & 45.0 & 65.0 & 30.0 & 35.0 & 35.7894737 \\
        & No Comp     & 40.0 & 5.0 & 25.0 & 40.0 & 35.0 & 40.0 & \textbf{75.0} & 35.0 & 15.0 & 32.3684211 \\
        & Zeroshot  & 50.0 & 10.0 & 15.0 & 40.0 & 45.0 & 40.0 & 60.0 & \textbf{65.0} & 30.0 & 35.2631579 \\
        \midrule
        \multirow{3}{*}{\shortstack[l]{Gemini 2.5 Pro \\ (Preview)}}
        & Hint      & 40.0 & 10.0 & 25.0 & 75.0 & \textbf{65.0} & \textbf{55.0} & 30.0 & 30.0 & \textbf{85.0} & 44.2105263 \\
        & No Comp     & \textbf{55.0} & \textbf{15.0} & 30.0 & \textbf{80.0} & 60.0 & 40.0 & 35.0 & 30.0 & \textbf{85.0} & \textbf{45.8} \\
        & Zeroshot  & \textbf{55.0} & \textbf{15.0} & 40.0 & 70.0 & \textbf{65.0} & \textbf{55.0} & 30.0 & 35.0 & \textbf{85.0} & 44.4736842 \\
        \bottomrule
        \end{tabular}
        \end{adjustbox}
    \end{subtable}
    \label{tab:llm_accuracy_percent_split_thinking} 
\end{table}

\begin{table}[!ht]
    \centering 
    \sisetup{round-mode=places, round-precision=1, table-auto-round} 
    \small 

    \caption{LLM token usage for reasoning models. The first subtable shows the token usage for the first 10 problem sets, and the second subtable shows the token usage for the remaining 9 problem sets and the overall average (AVG). The full result, including non-reasoning models, is in the Appendix.}
    
    \begin{subtable}{\linewidth}
        \centering
        \label{subtab:token_usage_problems_1_10_thinking}
        \begin{adjustbox}{max width=\textwidth, center}
        \begin{tabular}{@{}ll *{10}{S[table-format=5.1]}@{}}
        \toprule
        \textbf{Model} & \textbf{Prompt} &
        \colhead{RGPow} & \colhead{SqGrd} & \colhead{QSpin} & \colhead{CFTOp} & \colhead{3DBF} &
        \colhead{GSDGen} & \colhead{WrdH} & \colhead{Ferm} & \colhead{DynCon} & \colhead{3DEF} \\
        \cmidrule(lr){3-12}

        \multirow{3}{*}{DeepSeek R1}
        & Hint      & 7748.4 & 6085.45 & 7632 & 6711.2 & 11371.5 & 12606.65 & 10490.3 & 9499.3 & 8516.8 & 9784.65 \\
        & No Comp     & 6183.85 & 3830.25 & 8215.8 & 7438.85 & 10062 & 10215.3 & 10626.05 & 9688.75 & 9623.25 & 9504.5 \\
        & Zeroshot  & 7317.75 & 3968.15 & 6483.65 & 7647.75 & 11879.85 & 12193.4 & 10163.15 & 9927 & 8581.3 & 9954.55 \\
        \midrule
        \multirow{3}{*}{\shortstack[l]{Claude 3.7 \\ Sonnet \\ (Thinking)}}
        & Hint      & 17735.05 & 15298.55 & 18616.25 & 18987.75 & 19144.3 & 20767.25 & 20272.15 & 21516.2 & 13818.15 & 19036.15 \\
        & No Comp     & 16001.6 & 15081.4 & 17857.75 & 17108.1 & 20267.2 & 20318.35 & 22182.9 & 20201.05 & 15469.75 & 18977.35 \\
        & Zeroshot  & 15936 & 17851 & 18357.8 & 15954.55 & 21768.8 & 20952.5 & 20672.05 & 18951.2 & 14550.9 & 15786.8 \\
        \midrule
        \multirow{3}{*}{O4-Mini-High}
        & Hint      & 5345 & 6086.45 & 3813.25 & 1545.5 & 13739.4 & 10854.55 & 2566.25 & 8714.3 & 4469.25 & 8173.65 \\
        & No Comp     & 4575.9 & 3702.05 & 3336.9 & 1624.75 & 15863.4 & 9539.9 & 2130.9 & 8756.7 & 4529.95 & 8640.45 \\
        & Zeroshot  & 6536.9 & 5555.7 & 4028.6 & 2661.45 & 13835.55 & 13012.35 & 3477.75 & 8909.45 & 5576.65 & 11134.45 \\
        \midrule
        \multirow{3}{*}{\shortstack[l]{Gemini 2.5 Pro \\ (Preview)}}
        & Hint      & 20051.15 & 21261.65 & 22041.2 & 19477.05 & 26953.95 & 26402.8 & 15640.6 & 26444.4 & 22526 & 21781.75 \\
        & No Comp     & 17567.4 & 18813.4 & 21876.45 & 17009.9 & 26479.1 & 24327.95 & 16612.9 & 23438.2 & 22199.9 & 22420.25 \\
        & Zeroshot  & 20182.7 & 20409.15 & 20523.45 & 19394.35 & 26385.5 & 26236.15 & 16700.3 & 23850.2 & 21939.15 & 21351.4 \\
        \bottomrule
        \end{tabular}
        \end{adjustbox}
    \end{subtable}

    \vspace{0.3em}
    \begin{subtable}{\linewidth}
        \centering
        \label{subtab:token_usage_problems_11_avg_thinking}
        \begin{adjustbox}{max width=\textwidth, center}
        \begin{tabular}{@{}ll *{10}{S[table-format=5.1]}@{}}
        \toprule
        \textbf{Model} & \textbf{Prompt} &
        \colhead{DimLS} & \colhead{GpEdg} & \colhead{GSDeg} & \colhead{Solub} & \colhead{2DEF} &
        \colhead{2DEFL} & \colhead{OthGrd} & \colhead{2DBF} & \colhead{InfRes} & \colhead{Avg} \\
        \cmidrule(lr){3-12}

        \multirow{3}{*}{DeepSeek R1}
        & Hint      & 6586.6 & 5079.4 & 11792.4 & 2220.35 & 8191.65 & 10516.3 & 3230.9 & 10219.9 & 10018.2 & 8331.681579 \\
        & No Comp     & 5322.9 & 5178.85 & 10077.75 & 2745.5 & 7489.6 & 9865 & 2512 & 10132.25 & 6919.35 & 7664.831579 \\
        & Zeroshot  & 5014.05 & 5365.7 & 11397.6 & 2220.3 & 7621.45 & 10445.95 & 3458.55 & 11323.9 & 8648.95 & 8084.894737 \\
        \midrule
        \multirow{3}{*}{\shortstack[l]{Claude 3.7 \\ Sonnet \\ (Thinking)}}
        & Hint      & 14052.55 & 17293.1 & 29709.8 & 8725.1 & 16925.4 & 25224 & 12846.45 & 19499.4 & 16712.75 & 18220.01842 \\
        & No Comp     & 14565.5 & 19226.3 & 22652.25 & 11181.5 & 17803.3 & 24000.95 & 13721.2 & 18717.15 & 14370.95 & 17879.18684 \\
        & Zeroshot  & 11997.8 & 17201.9 & 26542 & 10072.25 & 16644.4 & 23130.25 & 12099.85 & 20960.05 & 14012.4 & 17549.60526 \\
        \midrule
        \multirow{3}{*}{O4-Mini-High}
        & Hint      & 3911.7 & 3958.65 & 9079.05 & 1754.15 & 7551.6 & 14260.35 & 4648.6 & 12895.05 & 9460.25 & 6990.894737 \\
        & No Comp     & 3673.2 & 3792 & 6692.95 & 2427.75 & 6766.7 & 12709.05 & 4341.45 & 9266.6 & 9333.4 & 6405.473684 \\
        & Zeroshot  & 4134.4 & 5193.5 & 8080.3 & 1912.05 & 8318.4 & 13526.95 & 6436.95 & 12987.45 & 16460.9 & 7988.407895 \\
        \midrule
        \multirow{3}{*}{\shortstack[l]{Gemini 2.5 Pro \\ (Preview)}}
        & Hint      & 12844.7 & 18129.4 & 24595.35 & 8205.2 & 22265.6 & 24617.9 & 18649.2 & 28798.45 & 15979.4 & 20877.14474 \\
        & No Comp     & 12147.8 & 17456.95 & 24887.35 & 8073.3 & 21378.95 & 23503 & 19128.6 & 30710.35 & 17622.1 & 20297.57105 \\
        & Zeroshot  & 13888.15 & 18554.35 & 25820.85 & 8387.15 & 22079.65 & 24866.75 & 18291.35 & 29737.3 & 18759.45 & 20913.54474 \\
        \bottomrule
        \end{tabular}
        \end{adjustbox}
    \end{subtable}
    \vspace{\baselineskip}
    \label{tab:llm_token_usage_split_thinking} 
\end{table}

\subsection{Reasoning Prompt Analysis}

Figure \ref{fig:average_accuracy} shows that the ``Hint'' and ``No-Comp'' prompts do not significantly improve the LLM’s overall accuracy. One possible explanation is that the models already know when to invoke physical principles, and their primary errors arise from misapplying those principles rather than failing to use them. Nonetheless, there are instances in which the LLM either fails to employ these principles or cannot determine the correct way to apply them, even with the “Hint” prompt. Below, we highlight an example. See Appendix A for another example.

\begin{myexample}{2D Electric Field (2DEF)}{example2}
    There is a uniformly charged plane in space. The plane is a square, and the four corners are at $(x,y,z) = (\pm 1, \pm 1, 0)$. Which of the following locations in space have the x-direction electric field strength equal to the y-direction electric field strength in both sign and magnitude, i.e. $E_x=E_y$?

    a) $(x=0, y=0, z=1)$;
    b) $(x=0, y=1, z=1)$;
    c) $(x=0, y=-1, z=1)$;
    d) $(x=1, y=0, z=1)$;
    e) $(x=1, y=1, z=1)$;
    f) $(x=1, y=-1, z=1)$;
    g) $(x=-1, y=0, z=1)$;
    h) $(x=-1, y=1, z=1)$;
    i) $(x=-1, y=-1, z=1)$;
    j) $(x=0, y=0, z=-1)$;
    k) $(x=0, y=1, z=-1)$;
    l) $(x=0, y=-1, z=-1)$;
    m) $(x=1, y=0, z=-1)$;
    n) $(x=1, y=1, z=-1)$;
    o) $(x=1, y=-1, z=-1)$;
    p) $(x=-1, y=0, z=-1)$;
    q) $(x=-1, y=1, z=-1)$;
    r) $(x=-1, y=-1, z=-1)$;
    s) $(x=2, y=0, z=0)$;
    t) $(x=2, y=1, z=0)$;
    u) $(x=2, y=-1, z=0)$;
    v) $(x=2, y=2, z=0)$;
    w) $(x=2, y=-2, z=0)$;
    x) $(x=0, y=2, z=0)$;
    y) $(x=1, y=2, z=0)$;
    z) $(x=-1, y=2, z=0)$;
\end{myexample}
\begin{myanswer}{``Zero Shot'' prompt}{answer2}
The electric field components at a point $(x, y, z)$ due to a uniformly charged square plane with corners at $( \pm 1, \pm 1,0)$ are given by: $E_x(x, y, z)=$ $\int_{-1}^1 \int_{-1}^1 \frac{k \sigma\left(x-x^{\prime}\right)}{\left(\left(x-x^{\prime}\right)^2+\left(y-y^{\prime}\right)^2+z^2\right)^{3 / 2}} d x^{\prime} d y^{\prime}$ $E_y(x, y, z)=\int_{-1}^1 \int_{-1}^1 \frac{k \sigma\left(y-y^{\prime}\right)}{\left(\left(x-x^{\prime}\right)^2+\left(y-y^{\prime}\right)^2+z^2\right)^{3 / 2}} d x^{\prime} d y^{\prime}$ We are looking for points where $E_x(x, y, z)=E_y(x, y, z)$
......
The final answer is: \boxed{a,b}
\end{myanswer}
In the answer from Gemini-2.5 Pro with ``Zero Shot'' prompt (shown above, details in the middle are omitted), the LLM utilizes complicated 2D integral but arrives at a wrong answer, showcasing the incapability of applying the symmetry principle which can greatly simplify the problem.
For physicists, this is an easy problem: by symmetry, all locations with $x=y$ should be selected, and thus the correct answer is (a,e,i,j,n,r,v)

\subsection{Reasoning Models vs. Non-reasoning models}
While many LLMs understand physical principles and can apply them in problem-solving, few consistently use these principles correctly or comprehensively. In particular, most LLMs struggle with the principle of symmetry, frequently failing to identify all relevant symmetries within a given system and sometimes incorrectly assuming symmetries that do not actually exist.

However, capabilities vary significantly between reasoning and non-reasoning models. Our experiments indicate that reasoning models are generally more effective at correctly applying physical principles. In contrast, non-reasoning models, despite being familiar with these principles conceptually, often demonstrate poor performance in fully understanding and applying them.

In the following, we illustrate this difference by examining and comparing responses from a reasoning and a non-reasoning model using an example drawn from quantum mechanics (see Appendix~\ref{app:answers} for detailed answers).

\begin{myexample}{Quantum dynamics (DynCon)}{example3}
Consider a $L=100$ quantum spin chain prepared as the ground state of the Hamiltonian $H = -\sum_j X_{j=1}^{100} - 0.9\sum_{j=1}^{99} Z_j Z_{j+1}$. Time-evolve this state under $H(t) = \sum_{j=1}^{97} Y_j X_{j+1} X_{j+2} Y_{j+3}$ from $t=0$ to $t=100$. Which of the following is true in the final state?

    a) $\langle Z_{60} \rangle = 0$;
    b) $\langle Z_{39}Y_{40} \rangle = \langle Y_{90}Z_{91} \rangle$;
    c) $\langle Z_{39}X_{40} \rangle = \langle X_{61}Z_{62} \rangle$;
    d) None of above is true.
\end{myexample}

\textbf{Human Reasoning:} This question is easy for humans. Choices (a), (b), and (c) are all correct thanks to spin-flip, time-reversal, and reflection symmetries, respectively (note that choice (c) can also be justified by spin-flip symmetry, since both sides evaluate to zero).

\textbf{Reasoning models:} The reasoning model (OpenAI/o4-mini-high) correctly identifies spin-flip and reflection symmetries but fails to apply time-reversal symmetry. Even when prompted with a hint --- “Consider the inherent symmetries of the system; leveraging these can help solve the problem” --- it still overlooks time-reversal symmetry, suggesting this concept may be underrepresented in its training data. See Appendix for detailed answers of the LLM.


\textbf{Non-reasoning models:} The non-reasoning model (OpenAI/GPT4.1) likewise mentions spin-flip and reflection symmetries but omits choice (c), revealing only a superficial grasp of these principles --- (c) follows from either symmetry. When given the same hint, it does select (a), (b), and (c) and correctly applies spin-flip and reflection symmetries, but it incorrectly asserts that choice (b) relies on translational symmetry which is absent in the system. Like the reasoning model, it shows no awareness of time-reversal symmetry.

Overall, non-reasoning models demonstrate a shallow understanding of physical principles and often apply them inconsistently, showing a tendency to latch onto terminology without true comprehension.

\section{Limitations}
While the \textit{PhySense} benchmark offers a novel approach to evaluating principle-based physics reasoning in LLMs, its current iteration is exclusively text-based, omitting multi-modal reasoning, and its scope could be expanded to cover more physics subfields and problem complexities. Future enhancements should therefore aim to incorporate multi-modal problems, diversify physics topics and complexity, develop more nuanced evaluation metrics for reasoning quality.

\section{Conclusion}

We introduce \textit{PhySense}, a comprehensive, novel, human-curated principle-based physics reasoning benchmark for evaluating large language models on scientific problem-solving across diverse physics domains. \textit{PhySense} comprises 380 carefully designed problems spanning symmetry reasoning, dimensional analysis, renormalization-group analysis, topology, quantum dynamics, and more, together with three prompting strategies ("Zero shot", "Hint", and "No-computation"). Our extensive evaluation of seven state-of-the-art LLMs, including reasoning and non-reasoning models, reveals that while reasoning-focused LLMs outperform their non-reasoning counterparts, all models remain substantially below expert human performance. We observe consistent deficits in token efficiency, principled application of physical laws, and generalization across topics. Moreover, auxiliary prompting strategies (e.g., hints or “no-computation” directives) yield only marginal improvements, indicating the need for deeper integration of principle-based thinking to LLMs. For future directions, it will be important to try improving LLM's principle-based reasoning via supervised fine tuning or reinforcement learning. Our study provides valuable insights and guidance for developing LLMs with efficient, robust and interpretable principle-based reasoning, which are crucial for scientific collaborations and discoveries (see Appendix C for Broader Impacts).

\clearpage

\bibliographystyle{unsrtnat}
\bibliography{ref}


\clearpage

\appendix

\section{Analysis of Several Examples}

\subsection{An example of LLM failing to apply principles}\label{app:anotherexample}
For some problems, even if the LLM is forced to use principle, it fails to figure out the correct way to use it. In the following example, Gemini-2.5 Pro fails to find the correct symmetry of the system. In general, the LLMs we test have better performance with symmetry group of a square lattice, but for other cases like triangular or honeycomb lattices, the LLMs have difficulty finding out the symmetry group to consider (see below).
\begin{myexample}{2D Electric Field Lattice (2DEFL)}{example3}
    There are point charges on the infinite x-y plane in space. The coordinates of the charges are $\{(x,y,z)=m\cdot(\sqrt{3}/2,-1/2,0)+n\cdot(0,1,0) | (m+n)\not\equiv 0\mod 3, m,n \in Z\}$. i.e. forming a 2D infinite honeycomb lattice on the x-y plane. Each point has charge of equal amplitude. The charges with n being even integers are positive, and the charges with n being odd integers are negative.  Which of the following locations in space have the z-direction electric field strength $E_z$ being zero? 

    a) $(x=0, y=0, z=1)$;
    b) $(x=0, y=1/2, z=1)$;
    c) $(x=0, y=-1/2, z=1)$;
    d) $(x=0, y=1, z=1)$;
    e) $(x=0, y=-1, z=1)$;
    f) $(x=\sqrt{3}/2, y=0, z=1)$;
    g) $(x=\sqrt{3}/2, y=1/2, z=1)$;
    h) $(x=\sqrt{3}/2, y=-1/2, z=1)$;
    i) $(x=-\sqrt{3}/2, y=0, z=1)$;
    j) $(x=-\sqrt{3}/2, y=1/2, z=1)$;
    k) $(x=-\sqrt{3}/2, y=-1/2, z=1)$;
    l) $(x=\sqrt{3}/4, y=1/4, z=1)$;
    m) $(x=\sqrt{3}/4, y=-1/4, z=1)$;
    n) $(x=\sqrt{3}/4, y=3/4, z=1)$;
    o) $(x=\sqrt{3}/4, y=-3/4, z=1)$;
    p) $(x=-\sqrt{3}/4, y=1/4, z=1)$;
    q) $(x=-\sqrt{3}/4, y=-1/4, z=1)$;
    r) $(x=-\sqrt{3}/4, y=3/4, z=1)$;
    s) $(x=-\sqrt{3}/4, y=-3/4, z=1)$;
\end{myexample}
\begin{myanswer}{``Hint'' prompt}{answer2}
This problem requires a detailed analysis of the symmetries of the charge distribution. Given the complexity and the lack of an obvious symmetry that leads to 
$E_z = 0$ for the listed points without extensive calculation, and the peculiarity that some listed solutions are locations of charges themselves, it's possible that the problem requires advanced techniques or there is a specific property of this honeycomb lattice with alternating charges that is not immediately apparent ...... Due to the difficulty and potential complexity, i am unable to provide a justified answer. However, if this is from a known source, one might find existing literature on a similar problem. The final answer is \boxed{b,e,g,i,l,o,p,r}.
\end{myanswer}
The LLM does not even specify any symmetry to consider and the answer above is wrong. In fact, the correct use of symmetry is: $E_z=0$ can be fixed by a reflection + charge-conjugation combined symmetry, where the reflection plane includes a z-direction straight line, or by a rotation + charge-conjugation combined symmetry, where the rotation axis is a z-direction straight line. The charge configuration has such rotation + charge-conjugation combined symmetry at the middle points of the edges that connect a pair of positive and negative charges. Among the choices, those points are (f,i,o,r), which is our answer.

\subsection{Answers of Example 3}\label{app:answers}
\begin{myanswer}{o4-mini-high, ``Zero-shot'' prompt}{answer2}
Solution Outline
1. Symmetry 1: Global spin - flip in the X - basis

Define $\mathrm{Q}=\Pi_1{ }^{00} \mathrm{X}_{\mathrm{j}}$.
$-Q X_j Q=X_j, Q Z_j Q=-Z_j, Q Y_j Q=-Y_j$
- Both $\mathrm{H}_0=-\Sigma \mathrm{X}_{\mathrm{j}}-0.9 \Sigma \mathrm{Z}_{\mathrm{j}} \mathrm{Z}_{\mathrm{j}+1}$ and $\mathrm{H}_1=\Sigma \mathrm{Y}_{\mathrm{j}} \mathrm{X}_{\mathrm{j}_{+1}} \mathrm{X}_{\mathrm{j}_{+2}} \mathrm{Y}_{\mathrm{j}_{+3}}$ commute with Q
$\Rightarrow$ The state at all times is Q-symmetric $\Rightarrow\left\langle Z_j\right\rangle=0$
$\Rightarrow$ statement (a) is true.
2. Symmetry 2: Reflection about the chain center

Define $\mathrm{R}: \mathrm{j} \rightarrow 101-\mathrm{j}$. One checks $\mathrm{RH}_0 \mathrm{R}=\mathrm{H}_0$ and $\mathrm{RH}_1 \mathrm{R}=\mathrm{H}_1$. Thus the time - evolved state is R - symmetric.
Under R:
$Z_{39} X_{40} \rightarrow Z_{62} X_{61}=X_{61} Z_{62}$ (different - site Pauli commute)
$\Rightarrow\left\langle Z_{39} X_{40}\right\rangle=\left\langle X_{61} Z_{62}\right\rangle$
$\Rightarrow$ statement (c) is true.
3. Translation is not an exact symmetry of the open - chain $\mathrm{H}_0$ or $\mathrm{H}_1$, so (b) need not hold.

Final answer: $\boxed{\mathrm{a}, \mathrm{c}}$
\end{myanswer}

\begin{myanswer}{o4-mini-high, ``Hint'' prompt}{answer2}
Solution Outline:
1. Symmetry of the initial Hamiltonian $\mathrm{H}_0$ :
- $\mathrm{H}_0=-\Sigma_1{ }^{00} \mathrm{X}_{-} \mathrm{j}-0.9 \Sigma_1{ }^{99} \mathrm{Z}_{-} \mathrm{j} \mathrm{Z}_{-}\{\mathrm{j}+1\}$ commutes with the global spin - flip P= $\prod \mathrm{X} \_\mathrm{j}$.
- Hence its (unique) ground state $\left|\psi_0\right\rangle$ is a P - eigenstate and $\left\langle\mathrm{Z}_{-} \mathrm{j}\right\rangle_0=0$ for all j .
2. Time - evolution under $\mathrm{H}(\mathrm{t})=\Sigma_1{ }^{97} \mathrm{Y} \_\mathrm{jX} \_\{\mathrm{j}+1\} \mathrm{X} \_\{\mathrm{j}+2\} \mathrm{Y} \_\{\mathrm{j}+3\}$ :
- $\mathrm{H}(\mathrm{t})$ also commutes with $\mathrm{P} \Rightarrow\left\langle\mathrm{Z}_{-} \mathrm{j}\right\rangle_{-} \mathrm{t}$ remains zero $\Rightarrow$ statement $(\mathrm{a})$ is true.
3. Reflection symmetry $R(\mathrm{j} \rightarrow 101-\mathrm{j})$ :
- Both $\mathrm{H}_0$ and $\mathrm{H}(\mathrm{t})$ are invariant under R and $\left|\psi_0\right\rangle$ can be chosen R - symmetric.
- Hence $\langle A_i B_{\mathrm{i}+1}\rangle_t = \left\langle R A_i B_{\mathrm{i}+1} R \right\rangle_t = \left\langle A_{101-\mathrm{i}} B_{100-\mathrm{i}} \right\rangle_{t}$.
- For (c): i=39 gives (101-39,100-39)=(62,61), so $\left\langle Z_{39} X_{40}\right\rangle=\left\langle Z_{62} X_{61}\right\rangle=\left\langle X_{61} Z_{62}\right\rangle$ (operators on different sites commute) $\Rightarrow$ (c) is true.
4. Statement (b) involves $(39,40)$ vs $(90,91)$, which are neither related by $P$ nor by $R \Rightarrow$ false.

Final Answer: $\boxed{a,c}$
\end{myanswer}

\begin{myanswer}{o4-mini-high, ``No Comp'' prompt}{answer2}
    Step-by-step:
1. Define the global "spin - flip" operator

$$
G=\prod_{-}\{\mathrm{j}=1\}^{\wedge}\{100\} X_{-} \mathrm{j} .
$$

Note $GZ_jG = -Z_j$ and $GX_j G = +X_j$, so $G$ flips the sign of $Z$ and $Y$ and leaves $X$ invariant.
2. Check symmetries:
- $H_{initial}$ $=-\Sigma X_{-} \mathrm{j}-0.9 \Sigma Z_{-} \mathrm{j} Z_{-}\{\mathrm{j}+1\}$ commutes with G (the usual $Z_2$ symmetry of the transverse - field Ising model).
- $H_{drive}$ $=\sum\{j=1\}^{\wedge}\{97\} Y_{-} j X\{j+1\} X \_\{j+2\} Y_{-}\{j+3\}$ also commutes with $G$ (each term has two Y's). $\Rightarrow$ The time - evolved state remains a +1 eigenstate of $G$ for all $t$.
3. Consequences for expectation values:
- Any operator O with $\mathrm{GOG}=-\mathrm{O}$ has $\langle\mathrm{O}\rangle=0$ at all times.

In particular $Z_{-} j$ is G-odd $\Rightarrow\left\langle Z_{-} j\right\rangle=0 \Rightarrow(a)$ is true.
$-Z_{-} j X \_\{j+1\}$ is G-odd (one Z) $\Rightarrow\left\langle Z_{-} j X \_\{j+1\}\right\rangle=0$, and likewise $\left\langle X_{-} j Z_{-}\{j+1\}\right\rangle=0 \Rightarrow(c)$ holds $(0=0)$.
- $Z_{-} j Y_{-}\{j+1\}$ and $Y_{-} j Z_{-}\{j+1\}$ are each G-even, but there is no symmetry forcing $\left\langle Z_{-} j Y_{-}\{j+1\}\right\rangle=\left\langle Y_{-} k Z_{-}\{k+1\}\right\rangle$ for different string orderings $\Rightarrow$ (b) is not guaranteed.

Answer: $\boxed{a,c}$
\end{myanswer}

\begin{myanswer}{GPT-4.1, ``Zero-shot'' prompt}{answer2}
Let's analyze each statement:

Hamiltonians:
- Initial: $H_0=-\sum_{j=1}^{100} X_j-0.9 \sum_{j=1}^{99} Z_j Z_{j+1}$
(Transverse field Ising).
- Time-evolution: $H(t)=\sum_{j=1}^{97} Y_j X_{j+1} X_{j+2} Y_{j+3}$

Statement a: $\left\langle Z_{60}\right\rangle=0$ of the initial Hamiltonian has $\left\langle Z_j\right\rangle=0$ due to the symmetry. Time evolution with a symmetric Hamiltonian preserves this. So, this statement IS true.

Statement b: $\left\langle Z_{39} Y_{40}\right\rangle=\left\langle Y_{90} Z_{91}\right\rangle$

Statement c: $\left\langle Z_{39} X_{40}\right\rangle=\left\langle X_{61} Z_{62}\right\rangle$
Similarly, unless there is special symmetry (e.g., reflection through the center of the chain), these correlators will NOT be equal. Sites ( 39,40 ) and ( 61,62 ) are not mirror images either, so this is also FALSE.
Statement d: None of above is true.
Since (a) is true, this is FALSE.

Final Answer:
 $\boxed{a}$
\end{myanswer}

\begin{myanswer}{GPT-4.1, ``Hint'' prompt}{answer2}
Let's analyze the situation step-by-step using symmetries:
1. Initial Hamiltonian:

$$
H=-\sum_{j=1}^{100} X_j-0.9 \sum_{j=1}^{99} Z_j Z_{j+1}
$$

The ground state of this Hamiltonian is unique and respects all the symmetries of the Hamiltonian (e.g., parity symmetry).
2. Time-evolution Hamiltonian:

$$
H(t)=\sum_{j=1}^{97} Y_j X_{j+1} X_{j+2} Y_{j+3}
$$

This Hamiltonian is uniform and translation-invariant by 1 site. Let's check for parity symmetries and other symmetries.
a) $\left\langle Z_{60}\right\rangle=0$

Let's check if the evolved state is invariant under some symmetry that would force this expectation value to vanish.
- The time evolution Hamiltonian is even in $Y$ and $X$ but is odd under $Z \rightarrow-Z$ (since it contains only $Y$ and $X$ operators).
- The initial state is unique and symmetric; under the global spin-flip $Z \rightarrow-Z$, the initial state's expectation of any single $Z_j$ vanishes, i.e., $\left\langle Z_j\right\rangle=0$.
- The time-evolution Hamiltonian does not break this symmetry, so for all $t,\left\langle Z_j\right\rangle=0$.

Thus, statement a) is TRUE.
b) $\left\langle Z_{39} Y_{40}\right\rangle=\left\langle Y_{90} Z_{91}\right\rangle$
- The time-evolution Hamiltonian is translation-invariant.
- The initial state is translation-invariant.
- However, after time evolution, the system is still translation-invariant, so for any displacement $k,\left\langle Z_j Y_{j+1}\right\rangle$ should be the same for all valid $j$.
- However, notice the indices: $(39,40)$ and $(90,91)$. Both are in the bulk (well away from the boundaries), where translation invariance definitely holds.

So, statement b) is TRUE.
c) $\left\langle Z_{39} X_{40}\right\rangle=\left\langle X_{61} Z_{62}\right\rangle$
- Similarly, translation invariance applies.
- However, under certain reflection symmetries or if the operator is odd under any symmetry not respected by the Hamiltonian, the mean value could vanish or be related differently.
- But since both correlators are related by exactly the same relative placement, and the evolution/initial state are symmetric under translation, these expectation values should be equal.

Thus, statement c) is TRUE.
d) None of above is true.

At least one statement is true, so d) is FALSE.
Final Answer
$\boxed{a,b,c}$
\end{myanswer}

\begin{myanswer}{GPT-4.1, ``No comp'' prompt}{answer2}
Let's analyze each statement step-by-step:

Hamiltonians:
- Ground state prepared with:
$H=-\sum_{j=1}^{100} X_j-0.9 \sum_{j=1}^{99} Z_j Z_{j+1}$
- Time evolution under:
$H(t)=\sum_{j=1}^{97} Y_j X_{j+1} X_{j+2} Y_{j+3}$
- Time range: $t=0$ to $t=100$, so this is a long time evolution, likely causing full thermalization or strong scrambling.
(a) $\left\langle Z_{60}\right\rangle=0$ ?
- The evolution Hamiltonian does not conserve $Z_{60}$.
- Quantum thermalization in such a system would drive a single-site expectation like $\left\langle Z_{60}\right\rangle$ to zero in the absence of conserved quantities or symmetries.
- The initial state is even under spin flip, the evolution is also even under global Y flip, but there is no protection for $Z$-no field acts to bias it.

So (a) is likely true.

(b) $\left\langle Z_{39} Y_{40}\right\rangle=\left\langle Y_{90} Z_{91}\right\rangle$ ? this is crucial).
- The Hamiltonians are translationally invariant in the bulk, but 39-40 and 90-91 are separated by 51 sites, so are not related by any symmetry.
- No argument for equality of those two-point functions by symmetry.

(b) is not necessarily true.

(c) $\left\langle Z_{39} X_{40}\right\rangle=\left\langle X_{61} Z_{62}\right\rangle$ ?

Look at the operators:
- Sites $(39,40)$ and $(61,62)$ are separated by 22 sites. Again, in an open chain, these are not related by any symmetry.
- No evident symmetry or conservation relates these two correlators.

(c) is not necessarily true.

(d) None of the above is true?

Since (a) is true (generic local observable under thermalizing evolution approaches zero expectation in the absence of conservation laws), (d) is not true.

Final Answer:
$\boxed{a}$
\end{myanswer}
\clearpage
\section{Full Experiment Results}
\begin{table}[ht]
    \centering
    \sisetup{round-mode=places, round-precision=1, table-auto-round} 
    \small
    \caption{LLM accuracy scores (as percentages). The first table shows accuracy for the first 10 problem sets, and the second table shows accuracy for the remaining 9 problem sets and the overall average (AVG).}
    \begin{subtable}{\linewidth}
        \centering
        \label{subtab:accuracy_percent_problems_1_10}
        \begin{adjustbox}{max width=\textwidth, center}
        \begin{tabular}{@{}ll *{10}{S[table-format=3.1, table-space-text-post=\%]}@{}}
        \toprule
        \textbf{Model} & \textbf{Prompt} &
        \colhead{RGPow} & \colhead{SqGrd} & \colhead{QSpin} & \colhead{CFTOp} & \colhead{3DBF} &
        \colhead{GSDGen} & \colhead{WrdH} & \colhead{Ferm} & \colhead{DynCon} & \colhead{3DEF} \\
        \cmidrule(lr){3-12}

        \multirow{3}{*}{\shortstack[l]{Claude 3.7 \\ Sonnet}}
        & Hint      & 15.0 & 10.0 & 15.0 & 20.0 & 10.0 & 0.0 & 10.0 & 10.0 & 5.0 & 30.0 \\
        & No Comp     & 10.0 & 5.0 & 10.0 & 10.0 & 15.0 & 0.0 & 20.0 & 25.0 & 5.0 & 20.0 \\
        & Zeroshot  & 15.0 & 5.0 & 5.0 & 15.0 & 15.0 & 0.0 & 0.0 & 15.0 & 10.0 & 25.0 \\
        \midrule
        \multirow{3}{*}{DeepSeek R1}
        & Hint      & \textbf{25.0} & 5.0 & 0.0 & 0.0 & 5.0 & 0.0 & 55.0 & 0.0 & 0.0 & 5.0 \\
        & No Comp     & 15.0 & 5.0 & 5.0 & 0.0 & 5.0 & 0.0 & 40.0 & 5.0 & 0.0 & 0.0 \\
        & Zeroshot  & 10.0 & 0.0 & 10.0 & 5.0 & 10.0 & 0.0 & 30.0 & 5.0 & 0.0 & 5.0 \\
        \midrule
        \multirow{3}{*}{\shortstack[l]{Claude 3.7 \\ Sonnet \\ (Thinking)}}
        & Hint      & 5.0 & 30.0 & 35.0 & 30.0 & 40.0 & 0.0 & 65.0 & 20.0 & 5.0 & 50.0 \\
        & No Comp     & 10.0 & 25.0 & 30.0 & 20.0 & 45.0 & 0.0 & 25.0 & 10.0 & 0.0 & 30.0 \\
        & Zeroshot  & 10.0 & 45.0 & 35.0 & \textbf{35.0} & 35.0 & 0.0 & 30.0 & 25.0 & 5.0 & 30.0 \\
        \midrule
        \multirow{3}{*}{\shortstack[l]{DeepSeek \\ Chat V3}}
        & Hint      & 10.0 & 0.0 & 30.0 & 5.0 & 15.0 & 0.0 & 15.0 & 15.0 & 15.0 & 10.0 \\
        & No Comp     & 20.0 & 0.0 & 25.0 & 20.0 & 15.0 & 0.0 & 35.0 & 25.0 & 15.0 & 10.0 \\
        & Zeroshot  & 10.0 & 0.0 & 30.0 & 15.0 & 15.0 & 0.0 & 10.0 & 25.0 & 20.0 & 20.0 \\
        \midrule
        \multirow{3}{*}{GPT-4.1}
        & Hint      & 10.0 & 10.0 & 15.0 & 25.0 & 30.0 & 5.0 & 75.0 & 25.0 & 10.0 & 20.0 \\
        & No Comp     & 0.0 & 0.0 & 25.0 & 20.0 & 20.0 & 0.0 & 30.0 & 25.0 & 10.0 & 35.0 \\
        & Zeroshot  & 0.0 & 0.0 & 10.0 & 10.0 & 30.0 & 0.0 & 55.0 & \textbf{30.0} & 10.0 & 25.0 \\
        \midrule
        \multirow{3}{*}{O4-Mini-High}
        & Hint      & 5.0 & 20.0 & 45.0 & 25.0 & 35.0 & 15.0 & 70.0 & 50.0 & 0.0 & 40.0 \\
        & No Comp     & \textbf{25.0} & 10.0 & 45.0 & 20.0 & 45.0 & 20.0 & 50.0 & 35.0 & 5.0 & 50.0 \\
        & Zeroshot  & 15.0 & 15.0 & 35.0 & 15.0 & 45.0 & 20.0 & 80.0 & 15.0 & 10.0 & 65.0 \\
        \midrule
        \multirow{3}{*}{\shortstack[l]{Gemini 2.5 Pro \\ (Preview)}}
        & Hint      & 10.0 & \textbf{50.0} & 65.0 & 25.0 &\textbf{50.0} & 5.0 & \textbf{100.0} & \textbf{30.0} & 25.0 & 65.0 \\
        & No Comp     & 20.0 & 40.0 & 65.0 & 25.0 & \textbf{50.0} & \textbf{25.0} & \textbf{100.0} & 25.0 & 20.0 & \textbf{70.0} \\
        & Zeroshot  & 10.0 & 35.0 & \textbf{70.0} & 25.0 & 40.0 & 15.0 & 95.0 & 25.0 & \textbf{30.0} & 50.0 \\
        \bottomrule
        \end{tabular}
        \end{adjustbox}
    \end{subtable}

    \vspace{0.3em} 

    \begin{subtable}{\linewidth}
        \centering
        \label{subtab:accuracy_percent_problems_11_avg}
        \begin{adjustbox}{max width=\textwidth, center}
        \begin{tabular}{@{}ll *{10}{S[table-format=3.1, table-space-text-post=\%]}@{}}
        \toprule
        \textbf{Model} & \textbf{Prompt} &
        \colhead{DimLS} & \colhead{GpEdg} & \colhead{GSDeg} & \colhead{Solub} & \colhead{2DEF} &
        \colhead{2DEFL} & \colhead{OthGrd} & \colhead{2DBF} & \colhead{InfRes} & \colhead{AVG} \\
        \cmidrule(lr){3-12}

        \multirow{3}{*}{\shortstack[l]{Claude 3.7 \\ Sonnet}}
        & Hint      & 15.0 & 10.0 & 25.0 & 70.0 & 10.0 & 5.0 & 55.0 & 5.0 & 25.0 & 18.1578947 \\
        & No Comp     & 25.0 & 10.0 & 25.0 & 50.0 & 15.0 & 10.0 & 40.0 & 5.0 & 20.0 & 16.8421053 \\
        & Zeroshot  & 25.0 & 0.0 & 20.0 & 55.0 & 15.0 & 10.0 & 50.0 & 0.0 & 25.0 & 16.0526316 \\
        \midrule
        \multirow{3}{*}{DeepSeek R1}
        & Hint      & 15.0 & 5.0 & 30.0 & 0.0 & 10.0 & 5.0 & 10.0 & 10.0 & 40.0 & 11.5789474 \\
        & No Comp     & 0.0 & 0.0 & 20.0 & 0.0 & 5.0 & 0.0 & 15.0 & 5.0 & 70.0 & 10.0000000 \\ 
        & Zeroshot  & 5.0 & 10.0 & 30.0 & 0.0 & 5.0 & 0.0 & 0.0 & 0.0 & 45.0 & 8.9473684 \\
        \midrule
        \multirow{3}{*}{\shortstack[l]{Claude 3.7 \\ Sonnet \\ (Thinking)}}
        & Hint      & 45.0 & 5.0 & 15.0 & 60.0 & 40.0 & 15.0 & 35.0 & 15.0 & 50.0 & 29.4736842 \\
        & No Comp     & 45.0 & 5.0 & 20.0 & 65.0 & 45.0 & 15.0 & 40.0 & 5.0 & 35.0 & 24.7368421 \\
        & Zeroshot  & 50.0 & 5.0 & 30.0 & 65.0 & 35.0 & 15.0 & 40.0 & 15.0 & 35.0 & 28.4210526 \\
        \midrule
        \multirow{3}{*}{\shortstack[l]{DeepSeek \\ Chat V3}}
        & Hint      & 25.0 & 10.0 & 10.0 & 50.0 & 0.0 & 10.0 & 45.0 & 0.0 & 50.0 & 16.5789474 \\
        & No Comp     & 15.0 & 5.0 & 15.0 & 40.0 & 15.0 & 0.0 & 25.0 & 5.0 & 35.0 & 16.8421053 \\
        & Zeroshot  & 30.0 & 10.0 & 20.0 & 25.0 & 5.0 & 0.0 & 35.0 & 5.0 & 35.0 & 16.3157895 \\
        \midrule
        \multirow{3}{*}{GPT-4.1}
        & Hint      & 35.0 & 10.0 & 20.0 & 45.0 & 10.0 & 10.0 & 50.0 & 0.0 & 30.0 & 22.8947368 \\
        & No Comp     & 30.0 & 5.0 & 10.0 & 40.0 & 10.0 & 10.0 & 50.0 & 5.0 & 30.0 & 18.6842105 \\
        & Zeroshot  & 35.0 & 0.0 & 10.0 & 35.0 & 15.0 & 5.0 & 25.0 & 0.0 & 10.0 & 16.0526316 \\
        \midrule
        \multirow{3}{*}{O4-Mini-High}
        & Hint      & \textbf{55.0} & 10.0 & \textbf{45.0} & 45.0 & 45.0 & 45.0 & 65.0 & 30.0 & 35.0 & 35.7894737 \\
        & No Comp     & 40.0 & 5.0 & 25.0 & 40.0 & 35.0 & 40.0 & \textbf{75.0} & 35.0 & 15.0 & 32.3684211 \\
        & Zeroshot  & 50.0 & 10.0 & 15.0 & 40.0 & 45.0 & 40.0 & 60.0 & \textbf{65.0} & 30.0 & 35.2631579 \\
        \midrule
        \multirow{3}{*}{\shortstack[l]{Gemini 2.5 Pro \\ (Preview)}}
        & Hint      & 40.0 & 10.0 & 25.0 & 75.0 & \textbf{65.0} & \textbf{55.0} & 30.0 & 30.0 & \textbf{85.0} & 44.2105263 \\
        & No Comp     & \textbf{55.0} & \textbf{15.0} & 30.0 & \textbf{80.0} & 60.0 & 40.0 & 35.0 & 30.0 & \textbf{85.0} & \textbf{45.8} \\
        & Zeroshot  & \textbf{55.0} & \textbf{15.0} & 40.0 & 70.0 & \textbf{65.0} & \textbf{55.0} & 30.0 & 35.0 & \textbf{85.0} & 44.4736842 \\
        \bottomrule
        \end{tabular}
        \end{adjustbox}
    \end{subtable}
    \label{full_acc} 
\end{table}
\begin{table}[ht]
    \centering 
    \sisetup{round-mode=places, round-precision=1, table-auto-round} 
    \small 

    \caption{LLM performance scores across problem sets (H: Hint, N: No Comp, Z: Zeroshot). Table (a) shows results for the first 10 problem sets, and Table (b) shows results for the remaining 9 problem sets and the overall average score for each model configuration.}
    
    \begin{subtable}{\linewidth}
        \centering
        \label{subtab:problems_1_10}
        \begin{adjustbox}{max width=\textwidth, center}
        \begin{tabular}{@{}ll *{10}{S[table-format=5.1]}@{}}
        \toprule
        \textbf{Model} & \textbf{Prompt} &
        \colhead{RGPow} & \colhead{SqGrd} & \colhead{QSpin} & \colhead{CFTOp} & \colhead{3DBF} &
        \colhead{GSDGen} & \colhead{WrdH} & \colhead{Ferm} & \colhead{DynCon} & \colhead{3DEF} \\
        \cmidrule(lr){3-12}

        \multirow{3}{*}{\shortstack[l]{Claude 3.7 \\ Sonnet}}
        & Hint      & 512.75 & 349.05 & 468.8 & 343.6 & 392.3 & 354.3 & 768.6 & 471.1 & 454.9 & 447.85 \\
        & No Comp     & 469.6 & 403.25 & 474.25 & 401.3 & 445.7 & 398.45 & 724.9 & 507.65 & 529.95 & 485.2 \\
        & Zeroshot  & 400.65 & 382.65 & 462.45 & 341.8 & 405.9 & 376 & 718.5 & 459.9 & 510.25 & 469.55 \\
        \midrule
        \multirow{3}{*}{DeepSeek R1}
        & Hint      & 7748.4 & 6085.45 & 7632 & 6711.2 & 11371.5 & 12606.65 & 10490.3 & 9499.3 & 8516.8 & 9784.65 \\
        & No Comp     & 6183.85 & 3830.25 & 8215.8 & 7438.85 & 10062 & 10215.3 & 10626.05 & 9688.75 & 9623.25 & 9504.5 \\
        & Zeroshot  & 7317.75 & 3968.15 & 6483.65 & 7647.75 & 11879.85 & 12193.4 & 10163.15 & 9927 & 8581.3 & 9954.55 \\
        \midrule
        \multirow{3}{*}{\shortstack[l]{Claude 3.7 \\ Sonnet \\ (Thinking)}}
        & Hint      & 17735.05 & 15298.55 & 18616.25 & 18987.75 & 19144.3 & 20767.25 & 20272.15 & 21516.2 & 13818.15 & 19036.15 \\
        & No Comp     & 16001.6 & 15081.4 & 17857.75 & 17108.1 & 20267.2 & 20318.35 & 22182.9 & 20201.05 & 15469.75 & 18977.35 \\
        & Zeroshot  & 15936 & 17851 & 18357.8 & 15954.55 & 21768.8 & 20952.5 & 20672.05 & 18951.2 & 14550.9 & 15786.8 \\
        \midrule
        \multirow{3}{*}{\shortstack[l]{DeepSeek \\ Chat V3}}
        & Hint      & 665.1 & 475.55 & 623.75 & 382.8 & 1064 & 665 & 1274.75 & 717.8 & 1048.75 & 1144.75 \\
        & No Comp     & 623.1 & 475.75 & 562.2 & 407.7 & 930.75 & 756.7 & 1325.35 & 668.3 & 1008.4 & 1437.7 \\
        & Zeroshot  & 627.45 & 670.15 & 716.3 & 411.25 & 850.1 & 604.2 & 1291.35 & 742.15 & 823.15 & 1489.7 \\
        \midrule
        \multirow{3}{*}{GPT-4.1}
        & Hint      & 633.6 & 678.4 & 671.2 & 480.4 & 869.55 & 1091.4 & 1300.4 & 715.6 & 709.05 & 1112.15 \\
        & No Comp     & 674.9 & 732.95 & 725.9 & 587.4 & 1058.25 & 1024.2 & 1188.85 & 839.8 & 913.45 & 1278.65 \\
        & Zeroshot  & 516.75 & 577.8 & 541.85 & 466.5 & 878.35 & 914.3 & 1234.75 & 684.2 & 678.1 & 1044.8 \\
        \midrule
        \multirow{3}{*}{O4-Mini-High}
        & Hint      & 5345 & 6086.45 & 3813.25 & 1545.5 & 13739.4 & 10854.55 & 2566.25 & 8714.3 & 4469.25 & 8173.65 \\
        & No Comp     & 4575.9 & 3702.05 & 3336.9 & 1624.75 & 15863.4 & 9539.9 & 2130.9 & 8756.7 & 4529.95 & 8640.45 \\
        & Zeroshot  & 6536.9 & 5555.7 & 4028.6 & 2661.45 & 13835.55 & 13012.35 & 3477.75 & 8909.45 & 5576.65 & 11134.45 \\
        \midrule
        \multirow{3}{*}{\shortstack[l]{Gemini 2.5 Pro \\ (Preview)}}
        & Hint      & 20051.15 & 21261.65 & 22041.2 & 19477.05 & 26953.95 & 26402.8 & 15640.6 & 26444.4 & 22526 & 21781.75 \\
        & No Comp     & 17567.4 & 18813.4 & 21876.45 & 17009.9 & 26479.1 & 24327.95 & 16612.9 & 23438.2 & 22199.9 & 22420.25 \\
        & Zeroshot  & 20182.7 & 20409.15 & 20523.45 & 19394.35 & 26385.5 & 26236.15 & 16700.3 & 23850.2 & 21939.15 & 21351.4 \\
        \bottomrule
        \end{tabular}
        \end{adjustbox}
    \end{subtable}

    \vspace{0.3em}
    \begin{subtable}{\linewidth}
        \centering
        \label{subtab:problems_11_avg}
        \begin{adjustbox}{max width=\textwidth, center}
        \begin{tabular}{@{}ll *{10}{S[table-format=5.1]}@{}}
        \toprule
        \textbf{Model} & \textbf{Prompt} &
        \colhead{DimLS} & \colhead{GpEdg} & \colhead{GSDeg} & \colhead{Solub} & \colhead{2DEF} &
        \colhead{2DEFL} & \colhead{OthGrd} & \colhead{2DBF} & \colhead{InfRes} & \colhead{Avg} \\
        \cmidrule(lr){3-12}

        \multirow{3}{*}{\shortstack[l]{Claude 3.7 \\ Sonnet}}
        & Hint      & 528.75 & 435.2 & 404.05 & 334.6 & 374.5 & 368.05 & 351.75 & 404.45 & 340.85 & 426.6026316 \\
        & No Comp     & 617.75 & 462.3 & 397.9 & 392.15 & 427.1 & 422.35 & 422.5 & 403.8 & 357.65 & 460.1973684 \\
        & Zeroshot  & 509.7 & 426.3 & 351.05 & 316.9 & 400.8 & 375.9 & 363.9 & 382.3 & 322.7 & 419.8526316 \\
        \midrule
        \multirow{3}{*}{DeepSeek R1}
        & Hint      & 6586.6 & 5079.4 & 11792.4 & 2220.35 & 8191.65 & 10516.3 & 3230.9 & 10219.9 & 10018.2 & 8331.681579 \\
        & No Comp     & 5322.9 & 5178.85 & 10077.75 & 2745.5 & 7489.6 & 9865 & 2512 & 10132.25 & 6919.35 & 7664.831579 \\
        & Zeroshot  & 5014.05 & 5365.7 & 11397.6 & 2220.3 & 7621.45 & 10445.95 & 3458.55 & 11323.9 & 8648.95 & 8084.894737 \\
        \midrule
        \multirow{3}{*}{\shortstack[l]{Claude 3.7 \\ Sonnet \\ (Thinking)}}
        & Hint      & 14052.55 & 17293.1 & 29709.8 & 8725.1 & 16925.4 & 25224 & 12846.45 & 19499.4 & 16712.75 & 18220.01842 \\
        & No Comp     & 14565.5 & 19226.3 & 22652.25 & 11181.5 & 17803.3 & 24000.95 & 13721.2 & 18717.15 & 14370.95 & 17879.18684 \\
        & Zeroshot  & 11997.8 & 17201.9 & 26542 & 10072.25 & 16644.4 & 23130.25 & 12099.85 & 20960.05 & 14012.4 & 17549.60526 \\
        \midrule
        \multirow{3}{*}{\shortstack[l]{DeepSeek \\ Chat V3}}
        & Hint      & 1140.35 & 482.6 & 928 & 369.7 & 975.25 & 832.85 & 625.15 & 866.8 & 863.85 & 797.2 \\
        & No Comp     & 955.85 & 433.75 & 1094.7 & 336.95 & 714.05 & 744.55 & 541.3 & 827.2 & 770.15 & 769.1815789 \\
        & Zeroshot  & 987.7 & 456.35 & 729.3 & 292.7 & 839.4 & 831.5 & 597.05 & 839.25 & 424.35 & 748.6 \\
        \midrule
        \multirow{3}{*}{GPT-4.1}
        & Hint      & 1340.55 & 467.9 & 1269.8 & 262.1 & 692.75 & 941.3 & 548.5 & 676.95 & 348.25 & 779.4657895 \\
        & No Comp     & 1378.8 & 646.65 & 880.8 & 305.55 & 830.95 & 1144.35 & 594.05 & 833.65 & 438.7 & 846.2026316 \\
        & Zeroshot  & 1209.9 & 482.25 & 686 & 192.95 & 643.8 & 922.7 & 432.75 & 751.95 & 352.75 & 695.3921053 \\
        \midrule
        \multirow{3}{*}{O4-Mini-High}
        & Hint      & 3911.7 & 3958.65 & 9079.05 & 1754.15 & 7551.6 & 14260.35 & 4648.6 & 12895.05 & 9460.25 & 6990.894737 \\
        & No Comp     & 3673.2 & 3792 & 6692.95 & 2427.75 & 6766.7 & 12709.05 & 4341.45 & 9266.6 & 9333.4 & 6405.473684 \\
        & Zeroshot  & 4134.4 & 5193.5 & 8080.3 & 1912.05 & 8318.4 & 13526.95 & 6436.95 & 12987.45 & 16460.9 & 7988.407895 \\
        \midrule
        \multirow{3}{*}{\shortstack[l]{Gemini 2.5 Pro \\ (Preview)}}
        & Hint      & 12844.7 & 18129.4 & 24595.35 & 8205.2 & 22265.6 & 24617.9 & 18649.2 & 28798.45 & 15979.4 & 20877.14474 \\
        & No Comp     & 12147.8 & 17456.95 & 24887.35 & 8073.3 & 21378.95 & 23503 & 19128.6 & 30710.35 & 17622.1 & 20297.57105 \\
        & Zeroshot  & 13888.15 & 18554.35 & 25820.85 & 8387.15 & 22079.65 & 24866.75 & 18291.35 & 29737.3 & 18759.45 & 20913.54474 \\
        \bottomrule
        \end{tabular}
        \end{adjustbox}
    \end{subtable}
    \vspace{\baselineskip}
    \label{full_token} 
\end{table}

\clearpage
\section{Broader Impacts}
This work advances the efficiency, robustness and interpretability of scientific LLMs by highlighting their current limitations in principle-based physics reasoning and providing tools to diagnose and improve them. By promoting alignment with expert reasoning, it may lead to more trustworthy and efficient AI systems in scientific discovery, education, and decision-making. However, there is a potential risk that benchmarks like PhySense could be overfit by future models if it is used improperly, encouraging superficial pattern-matching rather than genuine conceptual understanding. Additionally, increased reliance on AI for scientific reasoning may inadvertently deskill human practitioners if not carefully integrated into collaborative workflows.
\section{Prompt Specification}

We construct the prompt in the following general way:

\begin{lstlisting}[basicstyle=\ttfamily\footnotesize]
"System Prompt" + "Problem Prompt"
\end{lstlisting}

For different types of problems, we add different system prompts at the beginning of the problem prompt:

\subsection{Numerical Problems}

The system prompt is:

\begin{lstlisting}[basicstyle=\ttfamily\footnotesize, breaklines=true]
For the given physics problem, please provide a concise and step-by-step solution.
Ensure that your final answer is placed within \boxed{}.
If the final answer contains fraction, \pi or \sqrt{2}, convert it to a decimal approximation and calculate the final numerical value,
using approximately 3.14159265359 for \pi and 1.41421356237 for \sqrt{2}.
The unit will be provided in the problem and should not be included in the answer.
The final answer should be expressed solely as a decimal number.
\end{lstlisting}

\subsection{Multiple Choice Problems}

The system prompt is:

\begin{lstlisting}[basicstyle=\ttfamily\footnotesize, breaklines=true]
For the given physics problem, please provide a concise and step-by-step solution.
Ensure that your final answer is placed within \boxed{}.
The choices you choose should be listed as a comma-separated list without any spaces.
The final answer should be expressed solely as a list of letters, such as a,b,c,d.
\end{lstlisting}

\subsection{Exact Match Problems}

The system prompt is:

\begin{lstlisting}[basicstyle=\ttfamily\footnotesize, breaklines=true]
For the given physics problem, please provide a concise and step-by-step solution.
Ensure that your final answer is placed within \boxed{}.
The final answer should be clearly expressed using the same notation as the problem statement.
\end{lstlisting}

\subsection{Problem Prompt Construction}

The problem prompt is constructed as follows:

\begin{lstlisting}[basicstyle=\ttfamily\footnotesize, breaklines=true]
"Question: " + problem text
\end{lstlisting}

Optionally, we can include the unit:

\begin{lstlisting}[basicstyle=\ttfamily\footnotesize, breaklines=true]
"The unit of the problem is: " + problem unit
\end{lstlisting}

And the problem prompt always ends with:

\begin{lstlisting}[basicstyle=\ttfamily\footnotesize, breaklines=true]
"Answer: "
\end{lstlisting}

\subsection{Principle Guidance}

For different principle guidance, we add the corresponding hints:

\subsubsection{Symmetry}

\begin{lstlisting}[basicstyle=\ttfamily\footnotesize, breaklines=true]
Consider the inherent symmetries of the system; leveraging these can help with solving the problem.
\end{lstlisting}

\subsubsection{RG Power Counting}

\begin{lstlisting}[basicstyle=\ttfamily\footnotesize, breaklines=true]
This problem is amenable to RG power counting; please leverage them when generating the solution.
\end{lstlisting}

\subsubsection{Dimensional Analysis, Limiting Case}

\begin{lstlisting}[basicstyle=\ttfamily\footnotesize, breaklines=true]
This problem can be approached by considering dimensional analysis and limiting cases; please utilize these concepts when solving.
\end{lstlisting}

\subsubsection{Dimensional Analysis}

\begin{lstlisting}[basicstyle=\ttfamily\footnotesize, breaklines=true]
The physical dimensions of the quantities are key; please utilize dimensional analysis to find a solution.
\end{lstlisting}

\subsubsection{Molecular Symmetry}

\begin{lstlisting}[basicstyle=\ttfamily\footnotesize, breaklines=true]
The solution can be found by analyzing the molecular symmetry; please apply these principles.
\end{lstlisting}

\subsubsection{Unitary Transformation}

\begin{lstlisting}[basicstyle=\ttfamily\footnotesize, breaklines=true]
A suitable unitary transformation can simplify or solve this problem; you can utilize it to find a solution.
\end{lstlisting}

\subsubsection{Topology}

\begin{lstlisting}[basicstyle=\ttfamily\footnotesize, breaklines=true]
Topological properties are central here; consider how these concepts can be applied to solve the problem.
\end{lstlisting}

\subsubsection{Equation of Motion}

\begin{lstlisting}[basicstyle=\ttfamily\footnotesize, breaklines=true]
The core of this problem lies in its dynamics; you can utilize the equations of motion to find a solution.
\end{lstlisting}

\subsection{No-Calculation Guidance}

If we want to add no-calculation guidance, we add the following to the problem prompt:

\begin{lstlisting}[basicstyle=\ttfamily\footnotesize, breaklines=true]
Please solve this problem symbolically without performing numerical computations.
Try to avoid complex calculation including integrals or derivatives, and focus on physics analysis.
\end{lstlisting}

\end{document}